\documentclass[pmlr,twocolumn,10pt]{jmlr} % W&CP article

\mlhtrack{proceedings}

\newif\iffinal
\finaltrue  % toggle this flag for camera-ready version after acceptance

\iffinal
    \ifmlhneedspmlr
      \jmlrvolume{346}
      \jmlryear{2026}
    \fi
    \ifmlhfindings \jmlrproceedings{}{ML4H 2026 - Findings Track}\fi
    \ifmlhdemo     \jmlrproceedings{}{ML4H 2026 - Demo Track}\fi
    \jmlrworkshop{Machine Learning for Health (ML4H) 2026}
\else
    \jmlrproceedings{}{Submitted to ML4H 2026: \mlhtrackname}
    \jmlrworkshop{Machine Learning for Health (ML4H) 2026}
\fi

\newif\ifshowrebuttalcolor

\showrebuttalcolorfalse

\usepackage{enumitem}
\usepackage{fancyvrb}
\usepackage{booktabs}
\usepackage{longtable}% for long tables
\usepackage{algorithm}%
\usepackage{algorithmicx}%
\usepackage{algpseudocode}%
\usepackage{dsfont}
\usepackage{amssymb}
\usepackage{makecell}
\usepackage{wrapfig}
\usepackage{adjustbox}
\usepackage[breakable]{tcolorbox}
\usepackage{xcolor}
\usepackage{array}
\usepackage{multirow}
\usepackage{multicol}
\usepackage{enumitem}
\usepackage{placeins}
\usepackage{siunitx}
\usepackage{amsmath}

\usepackage[switch]{lineno}

\theorembodyfont{\upshape}
\theoremheaderfont{\scshape}
\theorempostheader{:}
\theoremsep{\newline}

\title[Anchoring Clinical Events in Time]{Anchoring Clinical Events in Time: UID-Preserving Multimodal Reconstruction and Source-Grounded Adjudication}

\author[Kumar et al.]{%
  \Name{Sayantan Kumar \nametag{$^{1}$}}
  \and
  \Name{Nicolas Grimaldi\nametag{$^{1}$}}
  \and
  \Name{Jack Cummins\nametag{$^{2}$}}
  \and
  \Name{Jeremy C. Weiss\nametag{$^{1}$}}
  \\[3pt]
  \addr{$^{1}$National Library of Medicine, National Institutes of Health, USA}
  \\
  \addr{$^{2}$Princeton University, USA}
}

\definecolor{rebuttalblue}{RGB}{0, 102, 204}

\newcommand{\rebuttal}[1]{%
  \ifshowrebuttalcolor
    \textcolor{rebuttalblue}{#1}%
  \else
    #1%
  \fi
}

\begin{document}

% Adjust equation spacing globally

\maketitle

\ifmlhdemo\else
\begin{abstract}
\rebuttal{Interpreting a patient's hospital course depends on when symptoms developed, tests were performed, and treatments began. Discharge summaries often leave this chronology unclear, while structured electronic health record (EHR) tables timestamp only some of these events. We reconstruct clinical timelines by retrieving structured evidence for narrative events and jointly revising their estimated times. Each extracted occurrence carries a persistent unique identifier (UID) linked to its source span, keeping repeated mentions distinct throughout reconstruction. We also introduce GAVEL, a large language model (LLM) judge that checks disagreements between two timelines against the note and structured record and cites evidence for its verdicts. We evaluated six open-weight models on 40 clinician-annotated discharge summaries. Our pipeline recovered more reference events than LLM-Timeline and TKW2 when each comparison used the same backbone. For GLM-5.2, adding structured evidence improved timestamp agreement with essentially unchanged event recovery; ablations showed that links to timestamped source rows contributed to this improvement. Results varied across the other models, with some timing gains accompanied by fewer recovered events. Both GAVEL judges, GLM-5.2 and MiMo-V2.6-Pro-RL, favored multimodal over text-only GLM-5.2 timelines, while neither clearly distinguished the DeepSeek V3.2 variants. Manual review upheld most sampled GAVEL findings. The source links allow reconstructed times to be checked, and GAVEL identifies disagreements for human review.}
\end{abstract}

\fi

\ifmlhneedsstatements
\paragraph*{Data and Code Availability}
Code is available at \url{https://github.com/SayantanKumar/uid_anchor_detection}. MIMIC-IV data used in this study are available to credentialed researchers through PhysioNet (\url{https://physionet.org/content/mimiciv/3.1/}) under the applicable data use requirements.

\paragraph*{Institutional Review Board (IRB)}
This secondary analysis used deidentified data and did not constitute human-subjects research.
\fi

\begin{figure*}[t]
    \centering
    \includegraphics[width=\textwidth]{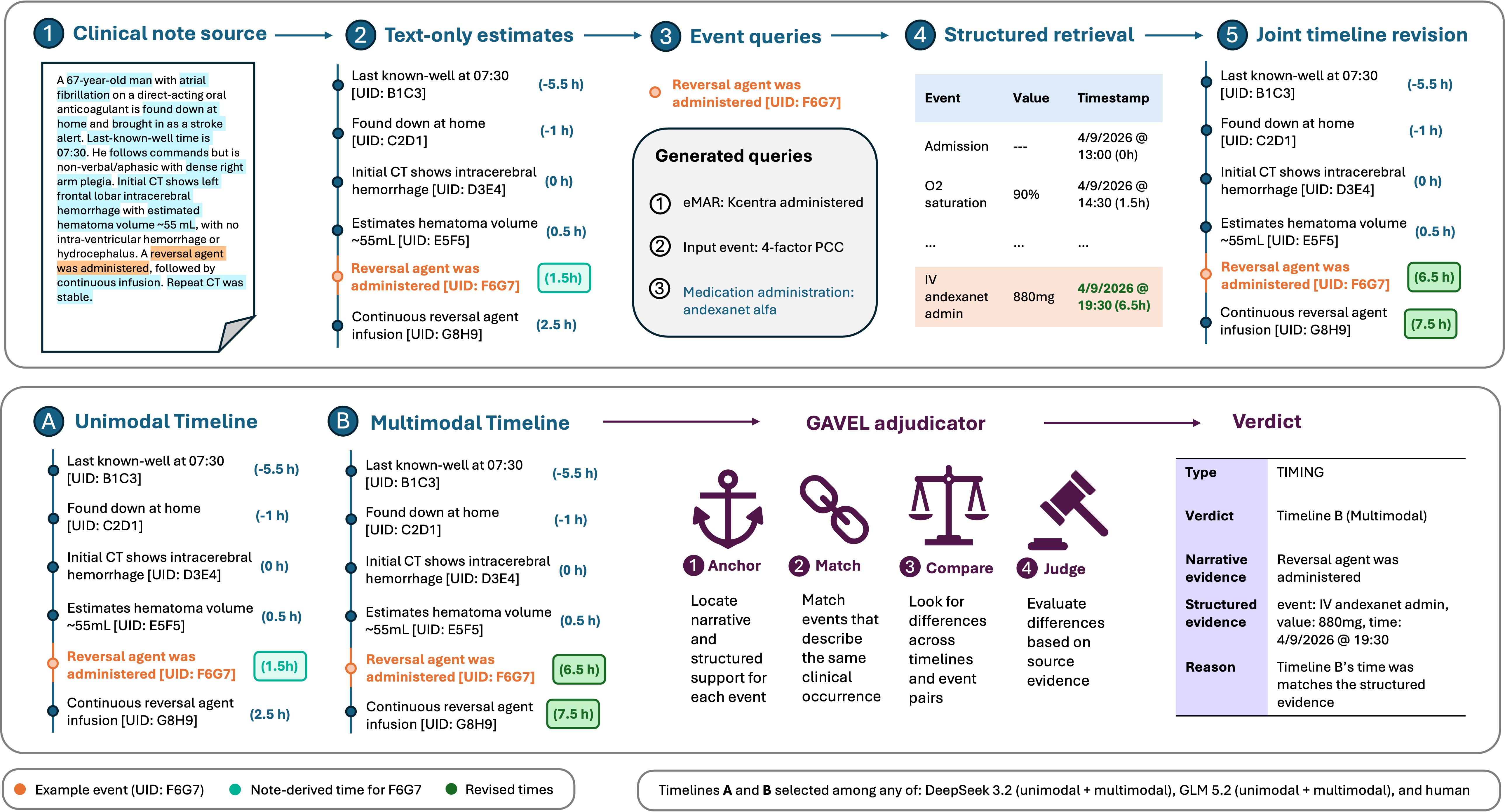}
    \vspace{-2mm}
    \caption{UID-preserving multimodal reconstruction and source-grounded adjudication in a synthetic intracerebral-hemorrhage case. \textbf{Top:} UID F6G7 links a narrative reversal event, its text-only estimate, retrieved administration evidence, and revised time. \textbf{Bottom:} GAVEL compares complete timelines against the narrative and structured record and returns an evidence-linked verdict.}
    \label{fig:overview}
    \vspace{-5mm}
\end{figure*}
%%%%%%%%%%%%%%%%%%%%%%%%%%%%%%%%%%%%%%%%%%%%%%

\vspace{-3mm}
\section{Introduction}
\label{sec:intro}
\vspace{-1mm}

In acute care, the order and timing of deterioration, diagnostic testing, treatment, and response determine how a patient's course is interpreted. \rebuttal{A shift of several hours can change the estimated interval between symptom onset and treatment, or whether a patient falls within a treatment window. We study \emph{multimodal clinical timeline reconstruction}: given a discharge summary and structured electronic health record (EHR) data from the same encounter, recover the clinical events described in the narrative and assign each a time relative to admission. Such timelines can support retrospective treatment-response analysis, temporal phenotyping, cohort construction, and trajectory modeling \citep{henry2022factors,kamran2024evaluation,noroozizadeh2023temporal}.}

This task requires two complementary but incomplete evidence sources. Structured EHR tables provide explicit timestamps for laboratory measurements, medications, procedures, and vital signs, but cover only part of the clinical course. On the contrary, unstructured discharge summaries on the other hand, preserve symptoms, progression, functional status, pertinent negatives, and clinical interpretation. However, they are written retrospectively and do not reflect the chronological sequence of the patient trajectory. \citep{moldwin2021empirical,seinen2025using,liu2022multimodal}. A structured timestamp may also denote ordering, collection, result, administration, or documentation rather than clinical onset. \rebuttal{Structured rows can therefore anchor selected narrative occurrences, while the note defines the clinical course to be reconstructed.}

\rebuttal{Earlier work progressed from temporal-relation extraction to absolute-time estimation and LLM-based timeline reconstruction \citep{sun2013evaluating,styler2014temporal,leeuwenberg2020towards,noroozizadeh2026reconstructing}. Structured EHR data can further reduce uncertainty in annotated event times \citep{frattallone2024using}. Our closest predecessor, \emph{Text Knows What, Tables Know When} (TKW2), calibrated a narrative event scaffold with retrieved structured rows, but passed events between stages as text descriptions \citep{kumar2026text}. After revision, the source occurrence corresponding to an event description can become ambiguous.
Descriptions of initial and repeat CT scans, recurrent weakness, or changes in medication status can use similar wording. When descriptions are merged, omitted, or rewritten, their link to the original occurrence can be lost. A plausible timestamp can then be attached to the wrong event.}

\rebuttal{Evaluation presents a related problem. Standard benchmarks align generated events to a clinician-authored timeline and measure event recovery, temporal ordering, and timestamp agreement \citep{wang2025large,noroozizadeh2026reconstructing}. However, they do not establish which account is supported when the candidate and reference disagree. The difference may reflect a model error, an annotation error, two defensible interpretations, or insufficient evidence. Checking the disagreement against the note and EHR tables can help determine whether the model or the reference annotation is wrong.}

\vspace{-1mm}
\paragraph{Contributions}
\rebuttal{Our contributions can be summarized as follows: \textbf{(i) UID-preserving reconstruction.} Each narrative occurrence receives a source-linked UID that persists through temporal inference, retrieval, source-row expansion, and joint revision. \textbf{(ii) Source-grounded adjudication.} GAVEL (\emph{Grounded Adjudication of Variations across Extracted timeLines}) checks two timelines against the note and structured record and returns an error type, verdict, and supporting evidence for each disagreement. \textbf{(iii) Evaluation across models and judges.} We evaluate six backbones on 40 discharge summaries using reference metrics and paired comparisons. We assess GAVEL through manual review and comparison of two judge families. Same-backbone comparisons show higher event recovery than LLM-Timeline and TKW2; temporal gains depend on the backbone.}

%%%%%%%%%%%%%%%%%%%%%%%%%%%%%%%%%%%%%
%%%%%%%%%%%%%%%%%%%%%%%%%%%%%%%%%%%%
\vspace{-3mm}
\section{Related Work}
\label{sec:related_work}
\vspace{-1mm}

\paragraph{Clinical temporal extraction and timeline reconstruction.}
Early clinical temporal NLP focused on event spans, temporal expressions, and pairwise relations such as \emph{before}, \emph{after}, \emph{overlap}, and \emph{containment} \citep{sun2013annotating,sun2013evaluating,styler2014temporal,lin2013medtime,lin2016multilayered}. Later methods organized these relations into patient timelines and assigned events explicit times or probabilistic temporal bounds \citep{nikfarjam2013towards,leeuwenberg2020towards}. Recent LLM-based systems reconstruct event--time trajectories from clinical narratives and use them for longitudinal analysis and forecasting \citep{wang2025large,noroozizadeh2026reconstructing,noroozizadeh2026forecasting}; ChemoTimelines evaluates patient-level treatment timeline extraction \citep{yao2025chemotimelines,zhang2025uwbionlp}. \rebuttal{TKW2 used a seven-step central/non-central scaffold with two structured-data calibration passes \citep{kumar2026text}. Our reconstruction pipeline starts with a a list of event mentions linked to their locations in the note and preserves it through evidence retrieval and joint revision. Appendix~\ref{apd:tkw2-comparison} gives the direct comparison.}

\vspace{-1mm}
\paragraph{Multimodal EHR grounding.}
Structured and narrative EHR data provide complementary information for phenotyping and patient representation \citep{liu2022multimodal,seinen2025using}. \rebuttal{For timeline reconstruction, the event sets overlap only partly: structured records timestamp some narrative events, while many symptoms, transitions, and contextual findings have no tabular counterpart.} \citet{frattallone2024using} showed that structured EHR data can refine the temporal bounds of annotated inpatient events. \rebuttal{Our pipeline links each retrieved source row to the narrative occurrence that prompted the search and preserve that link during timeline revision}

\vspace{-1mm}
\paragraph{LLM judges and source-grounded evaluation.}
Clinical timeline benchmarks typically align generated events to one clinician-authored reference and report event recovery and temporal agreement \citep{wang2025large,noroozizadeh2026reconstructing}. \rebuttal{General-purpose LLM judges assess response quality or preference \citep{zheng2023judging,chiang2024chatbotarena}, while FActScore verifies claims in one generation against supporting evidence \citep{min2023factscore}.} TIMER applies LLM-based evaluation to temporal reasoning over longitudinal EHRs and validates its scores against clinician rankings \citep{cui2025timer}. \rebuttal{GAVEL targets disagreements between two complete clinical timelines. It checks each disagreement against narrative and structured evidence. It can support one candidate, both, neither, or leave the disagreement unresolved. Its output identifies individual event, value, and timing discrepancies for review. Appendix~\ref{apd:related_work_details} provides the extended literature comparison.}

%%%%%%%%%%%%%%%%%%%%%%%%%%%%%%%%%%%%%
%%%%%%%%%%%%%%%%%%%%%%%%%%%%%%%%%%%%
\vspace{-3mm}
\section{Methods}
\label{sec:methods}
\vspace{-1mm}

Figure~\ref{fig:overview} summarizes the proposed framework. \rebuttal{The reconstruction pipeline extracts narrative occurrences, assigns persistent UIDs, estimates text-only times, retrieves structured evidence, and jointly revises the timeline. GAVEL is a separate, read-only evaluator and its findings do not feed back into reconstruction.}

\vspace{-2mm}
\subsection{Task and UID Representation}
\label{sec:task-uid}
\vspace{-1mm}

Let \(T\) denote a discharge summary and \(R=\{r_j\}_{j=1}^{M}\) the structured EHR rows from the same encounter, where \(r_j=(v_j,x_j,\tau_j)\) contains an event name, recorded value, and timestamp. We measure time in hours relative to hospital admission (\(t=0\)); if admission is unavailable, the earliest documented presentation defines the reference time. Negative values precede the reference and positive values follow it. 

For event occurrence \(i\), we define
\begingroup
\setlength{\abovedisplayskip}{2pt}
\setlength{\belowdisplayskip}{2pt}
\[
\begin{aligned}
z_i &= \bigl(u_i,m_i,p_i,\widehat{t}_i,I_i,k_i,C_i\bigr),\\
\widehat{\mathcal{S}} &= \{z_i\}_{i=1}^{N},
\end{aligned}
\]
\endgroup
where \(u_i\) is a case-local unique identifier (UID), \(m_i\) is the original mention, and \(p_i\) is its character span in \(T\). The remaining fields are the estimated time \(\widehat{t}_i\), plausible interval \(I_i=[\ell_i,h_i]\), indicator \(k_i\) of whether the narrative explicitly supports the time, and contextual UIDs \(C_i\) used in temporal reasoning.

\rebuttal{An event is a patient-specific clinical occurrence or state that can in principle be located in time. Each UID identifies one tagged source occurrence; separate mentions retain separate identifiers even when their wording is identical. UIDs track source mentions and fail to identify if different mentions describe the same underlying event. Independently generated event inventories cannot be aligned by UID alone.} These include but are not limited to patient-specific symptoms, diagnoses, findings, procedures, treatments, clinical states, outcomes, pertinent negatives, and transitions in care. Conjunctive findings are separated when they denote distinct occurrences, while modifiers that affect clinical or temporal interpretation are preserved. Full annotation rules are provided in Appendix~\ref{apd:event-inventory}.

\vspace{-2mm}
\subsection{UID-Preserving Multimodal Reconstruction}
\label{sec:uid-reconstruction}
\vspace{-1mm}

\paragraph{Event inventory and text-only initialization.}
\rebuttal{In the tagging stage, the LLM marks clinical event spans while preserving the surrounding note. The parser assigns a case-local UID to each tagged occurrence and links it to its source span (Appendix~\ref{apd:prompt-tagging}). A separate temporal-inference call to the same backbone receives the UID-tagged note and mention inventory without structured EHR evidence.}
For each UID, the model returns the original mention, a point estimate, temporal bounds, a binary \texttt{known} flag, and up to five contextual UIDs. The point estimate is the time of the event, relative to \(t=0\), of the most likely timeline. The bounds represent the earliest and latest times that could be supported by the narrative, while \texttt{known} distinguishes explicit timing from timing inferred through narrative order or clinical context. This output is both the text-only baseline and the starting point for multimodal revision (full prompt in Appendix~\ref{apd:prompt-text-only}).

\vspace{-1mm}
\paragraph{Query-conditioned structured-evidence retrieval.}
\rebuttal{We group structured rows by exact event type into patient-specific event-series summaries. For numeric series, we retain summary statistics and the values and timestamps of the first, last, minimum, and maximum observations. Categorical series are summarized by value frequencies. Short series retain all observations. Each summary stores the indices of its contributing raw rows.}

\rebuttal{For each UID, the query-generation stage produces up to three queries from the mention and its narrative context. A phrase such as ``reversal agent was administered'' can thus retrieve a named medication administration even when the note and table use different terminology (Appendix~\ref{apd:prompt-query-generation}). Qwen3-Embedding-8B retrieves the ten highest-ranked summaries per query, and Qwen3-Reranker-8B scores their relevance \citep{zhang2025qwen3embedding}. For each query, we retain up to three candidates with a reranker score of at least 0.05. We cap the total at 100 candidates per case.}

\rebuttal{Each retained summary is expanded through its stored row mapping to the original event names, values, and timestamps. The requesting UID and query remain attached to this evidence. A retrieved series may contain several occurrences of the same event or timestamps with different meanings. The revision prompt therefore instructs the model to distinguish order from administration and result time from specimen collection. Rows later than discharge plus 12 hours are excluded. Full summarization and filtering rules appear in Appendix~\ref{apd:event-inventory}.}

\vspace{-1mm}
\paragraph{Joint revision and provenance.}
\rebuttal{The revision stage receives the UID-tagged note, complete text-only timeline, UID-linked timestamped evidence, and admission and discharge times. It revises all event times together, using the sequence of presentation, testing, treatment, transfers, and outcomes as context. The revision prompt instructs the model to use a structured timestamp only when it supports the intended occurrence and timestamp type, and to retain the text-derived estimate otherwise (Appendix B.4) (Appendix~\ref{apd:prompt-joint-revision}).}

The model may revise point times, temporal bounds, and contextual links, but every valid output must contain each original UID and mention exactly once. It may not add, remove, merge, rename, or duplicate occurrences. The model may return up to three complete alternatives; Timeline~1 is the prespecified primary output. After validation, deterministic post-processing maps each UID back to its character span in the untagged note and retains the links from the occurrence to its queries, retrieved source rows, and final temporal assignment.

\vspace{-1mm}
\paragraph{\rebuttal{Worked example.}}
\rebuttal{In the synthetic case in Figure~\ref{fig:overview}, the note describes reversal-agent administration without stating when it occurred. The text-only stage places UID~\texttt{F6G7} at \(+1.5\) hours. A contextual query retrieves a medication event series; expansion exposes an andexanet alfa administration at 19:30, or \(+6.5\) hours from the 13:00 admission. Joint revision moves this occurrence to \(+6.5\) hours and places the subsequent infusion at \(+7.5\) hours, while retaining both UIDs. The row directly supports the administration time; the infusion time is inferred.}

\vspace{-2mm}
\subsection{GAVEL: Source-Grounded Timeline Adjudication}
\label{sec:gavel}
\vspace{-1mm}

\rebuttal{GAVEL compares two candidate timelines with the source record and reports disagreements without editing either candidate. It identifies differences in which events are included, their values and times, and whether they are repeated. It then checks each disagreement against the note and structured evidence.}

\rebuttal{The primary judge input contains the complete discharge summary, admission and discharge times, and two complete timelines serialized as event-and-time rows \emph{without UIDs}. GAVEL matches events by clinical meaning after inspecting both timelines. This allowed us to include the earlier clinician annotations, which lacked identifiers. A rerun with UID-bearing timelines is a separate sensitivity analysis (Appendix~\ref{apd:gavel_results}).}

\rebuttal{Both candidates are evaluated against the same encounter-level inventory of structured event-series summaries, capped at 900 summaries to fit the context budget. This inventory is independent of the evidence retrieved for either reconstruction. GAVEL’s structured evidence is limited to the summarized inventory. The prompt allows narrative evidence to support an event even when the event is absent from that inventory.}

\rebuttal{A frozen prompt (Appendix~\ref{apd:prompt-gavel}) performs four operations: \textbf{anchor} each event to narrative or structured support; \textbf{match} events by clinical meaning while retaining genuine recurrences; \textbf{compare} matched values and times; and \textbf{classify and judge} each discrepancy. Timing differences below the larger of three hours or 10\% of the event's distance from admission are suppressed. Before reporting a one-sided or duplicate event, the judge is instructed to scan the complete opposing timeline.}

GAVEL assigns one of six discrepancy types (\texttt{VALUE}, \texttt{TIMING}, \texttt{A\_ONLY}, \texttt{B\_ONLY}, \texttt{SHARED\_UNSUPPORTED}, or \texttt{DUPLICATE}) and one of five verdicts (\texttt{A}, \texttt{B}, \texttt{BOTH}, \texttt{NEITHER}, or \texttt{UNCLEAR}). Each finding contains the candidate rows, narrative and structured evidence, verdict, and a concise reason. \rebuttal{The protocol and output schema are specified in Appendix~\ref{apd:gavel_protocol}.}

%%%%%%%%%%%%%%%%%%%%%%%%%%%%%%%%%%
%%%%%%%%%%%%%%%%%%%%%%%%%%%%%%%%%

\vspace{-3mm}
\section{\rebuttal{Experimental Setup}}
\label{sec:experimental-setup}
\vspace{-1mm}

\paragraph{Cohort and clinician-authored timelines.}
We evaluate 40 discharge summaries spanning mixed critical-care presentations: 15 i2b2-derived summaries and 25 MIMIC-IV summaries, each linked to structured EHR data from the same encounter \citep{sun2013evaluating,johnson2023mimic,pollard_physionet_2026}.
\rebuttal{Three clinicians annotated the 40 cases (singly, non-overlapping), producing 9,147 reference events. The clinicians identified event spans in each discharge summary, inspected raw structured rows from the same hospitalization, and assigned times relative to admission. The note and structured rows came from the same hospitalization. No event-level links between narrative spans and EHR rows were supplied in advance.}

\rebuttal{The first 20 cases---15 i2b2-derived and five MIMIC-IV---came from the earlier annotation effort \citep{frattallone2024using} and were annotated without model suggestions. For the additional 20 MIMIC-IV cases, GLM-5.2 suggested candidate event and time anchors both structured and unstructured for the clinician to inspect. Event selection, evidence interpretation, and final timestamps remained the clinician's responsibility. The two case sets differed in both patients and data source. Comparisons between them therefore cannot isolate the effect of LLM assistance.}

\vspace{-1mm}
\paragraph{Models, baselines, and ablations.}
\rebuttal{We evaluate six open-weight backbones: GLM-5.2 FP8, DeepSeek V3.2, Qwen3.5-397B, GPT-OSS-120B, Mistral 3.5 Medium, and MiniMax M2.7. Within a run, the same backbone performs tagging, text-only inference, query generation, and revision, using stage-specific prompts without fine-tuning. Qwen3-Embedding-8B and Qwen3-Reranker-8B are fixed across runs. Inference is local; implementation and compute details appear in Appendix~\ref{apd:event-inventory}.}

\rebuttal{Table~\ref{tab:main-results} includes the single-step LLM-Timeline baseline with GLM-5.2 \citep{wang2025large} and TKW2 with DeepSeek V3.2 \citep{kumar2026text}. Both use the same evaluation cohort and matching threshold. These are cross-method comparisons; Appendix~\ref{apd:tkw2-comparison} details the pipeline differences, and Appendix~\ref{apd:prompt-single-step} gives the single-step prompt.}

\rebuttal{We compared the full pipeline with its text-only version and four component ablations. \emph{No UID} removes identifiers from joint revision while retaining the input event inventory, initial estimates, evidence, and backbone. \emph{No source-row linkage} supplies retrieved summaries without their linked timestamped rows. \emph{No query generation} retrieves using the mention string, and \emph{no reranker} omits reranking. Table~\ref{tab:best_ablation} reports all five paired GLM-5.2 comparisons; complete results for GLM-5.2, DeepSeek V3.2, and Qwen3.5-397B appear in Appendix~\ref{apd:baselines_ablations}.}

\begin{table*}[t]
\centering
\footnotesize
\setlength{\tabcolsep}{2.1pt}
\renewcommand{\arraystretch}{1.12}
\newcommand{\scoreci}[3]{#1 (#2--#3)}
\caption{\rebuttal{Reconstruction on 40 discharge summaries at matching threshold 0.1. Entries are estimates with 95\% marginal intervals from 200 case-bootstrap resamples. Prior methods are LLM-Timeline \citep{wang2025large} and TKW2 \citep{kumar2026text}; a dash denotes an unavailable modality.}}
\label{tab:main-results}
\begin{adjustbox}{max width=\textwidth}
\begin{tabular}{@{}lcccccc@{}}
\toprule
& \multicolumn{2}{c}{\textbf{Event match rate}}
& \multicolumn{2}{c}{\textbf{Concordance}}
& \multicolumn{2}{c}{\textbf{AULTC}} \\
\cmidrule(lr){2-3}\cmidrule(lr){4-5}\cmidrule(lr){6-7}
\textbf{Method / backbone} & \rebuttal{Text-only} & Multimodal & \rebuttal{Text-only} & Multimodal & \rebuttal{Text-only} & Multimodal \\
\midrule
\multicolumn{7}{@{}l}{\textit{Proposed UID-preserving method}} \\
\addlinespace[1pt]
GLM-5.2 & \scoreci{0.79}{0.75}{0.83} & \scoreci{0.79}{0.75}{0.83} & \scoreci{0.78}{0.76}{0.82} & \scoreci{0.80}{0.77}{0.84} & \scoreci{0.76}{0.72}{0.80} & \scoreci{0.77}{0.74}{0.81} \\
DeepSeek V3.2 & \scoreci{0.64}{0.59}{0.70} & \scoreci{0.62}{0.55}{0.67} & \scoreci{0.75}{0.72}{0.77} & \scoreci{0.76}{0.75}{0.82} & \scoreci{0.76}{0.72}{0.80} & \scoreci{0.75}{0.72}{0.80} \\
Qwen3.5-397B & \scoreci{0.75}{0.68}{0.83} & \scoreci{0.64}{0.53}{0.73} & \scoreci{0.76}{0.73}{0.81} & \scoreci{0.75}{0.72}{0.81} & \scoreci{0.74}{0.71}{0.78} & \scoreci{0.76}{0.74}{0.78} \\
GPT-OSS-120B & \scoreci{0.32}{0.28}{0.41} & \scoreci{0.32}{0.28}{0.42} & \scoreci{0.71}{0.67}{0.79} & \scoreci{0.72}{0.68}{0.79} & \scoreci{0.79}{0.72}{0.80} & \scoreci{0.79}{0.72}{0.82} \\
Mistral 3.5 Medium & \scoreci{0.75}{0.68}{0.81} & \scoreci{0.65}{0.53}{0.70} & \scoreci{0.75}{0.71}{0.81} & \scoreci{0.75}{0.71}{0.80} & \scoreci{0.78}{0.72}{0.80} & \scoreci{0.80}{0.73}{0.81} \\
MiniMax M2.7 & \scoreci{0.42}{0.38}{0.47} & \scoreci{0.42}{0.39}{0.45} & \scoreci{0.75}{0.70}{0.78} & \scoreci{0.74}{0.71}{0.77} & \scoreci{0.77}{0.73}{0.81} & \scoreci{0.77}{0.73}{0.80} \\
\midrule
\multicolumn{7}{@{}l}{\textit{Prior methods}} \\
\addlinespace[1pt]
LLM-Timeline (GLM-5.2) & \scoreci{0.55}{0.41}{0.60} & -- & \scoreci{0.74}{0.70}{0.76} & -- & \scoreci{0.80}{0.76}{0.81} & -- \\
TKW2 (DeepSeek V3.2) & \scoreci{0.48}{0.43}{0.56} & \scoreci{0.48}{0.43}{0.56} & \scoreci{0.75}{0.71}{0.80} & \scoreci{0.77}{0.73}{0.81} & \scoreci{0.76}{0.72}{0.79} & \scoreci{0.77}{0.73}{0.81} \\
\bottomrule
\end{tabular}
\end{adjustbox}
\vspace{-3mm}
\end{table*}

\vspace{-1mm}
\paragraph{\rebuttal{Evaluation of reconstructed timelines.}}
\rebuttal{Predicted and reference events are aligned one-to-one by recursively selecting the remaining pair with the smallest PubMedBERT cosine distance, using a primary threshold of 0.1. Event match rate measures reference-event recovery. Temporal concordance measures ordering agreement, and Area Under the Log-Time CDF (AULTC) summarizes absolute timestamp errors on a logarithmic time scale. Both temporal metrics use only matched events. AULTC uses the same five-year limit (43,800 hours) across methods, ablations, thresholds, and bootstrap replicates.}

\rebuttal{Confidence intervals use 200 case-level bootstrap resamples. For paired comparisons, we draw 10,000 case-level bootstrap samples from cases with valid outputs in both conditions. Each replicate uses the same sampled cases for both conditions and recomputes the metric difference. We report the 2.5th and 97.5th percentiles. Appendix~\ref{apd:tts_evaluation} details matching-threshold, order-permutation, and exact-duplicate sensitivity tests (Appendix~\ref{apd:event-inventory}).}

\vspace{-1mm}
\paragraph{\rebuttal{GAVEL evaluation.}}
\rebuttal{GAVEL compares five sources: the clinician-authored timeline and the text-only and multimodal timelines from GLM-5.2 and DeepSeek V3.2. The primary judge is GLM-5.2, with source identities concealed behind randomized labels A and B. For each decisive finding, we award points to the timeline favored by the judge. The error category determines the number of points. We fit Bradley–Terry ratings to the resulting per-case point shares. Direct within-backbone comparisons report the multimodal timeline's share of weighted points against its text-only counterpart, with 50\% denoting parity. Intervals use 2,000 case-level bootstrap resamples. Scoring rules and treatment of draws appear in Appendix~\ref{apd:gavel_protocol}.}

\rebuttal{We repeat the five-source analysis with MiMo-V2.6-Pro-RL, a different model family not used for reconstruction or reference annotation, keeping the prompt, evidence, weights, and scoring procedure fixed. Separate GLM-5.2 sensitivity runs reverse A/B positions and add UIDs to the candidate rows. Each analysis includes comparisons with valid judge outputs. The sets of valid comparisons differ between GLM-5.2 and MiMo.}

\vspace{-1mm}
\paragraph{\rebuttal{Manual review of GAVEL findings.}}
\rebuttal{The primary reviewer assessed 150 findings sampled from 10 cases. Sampling was restricted to the 36 cases with valid judgments for all three pairings among the clinician-authored, GLM-5.2 multimodal, and DeepSeek V3.2 multimodal timelines. Fifty findings were sampled per pairing, with uncommon types oversampled. A second reviewer independently scored one balanced packet of 50 findings. Reviewers saw GAVEL's verdict and recorded whether the reported discrepancy and verdict were justified. We weight each review stratum by its proportion among eligible GAVEL findings, with intervals from 20,000 stratified bootstrap resamples. Manual review was limited to findings reported by the primary GLM-5.2 judge, leaving missed-discrepancy recall unmeasured. Sampling details appear in Appendix~\ref{apd:gavel_protocol}.}

\vspace{-3mm}
\section{Results}
\label{sec:results}
\vspace{-1mm}

% Source: glm52_uid_ablation_paired_diff.csv; all comparisons use 40 cases.
\begin{table*}[t]
\centering
\footnotesize
\setlength{\tabcolsep}{5pt}
\renewcommand{\arraystretch}{1.08}
\caption{\rebuttal{Paired GLM-5.2 differences (full multimodal minus comparison) at matching threshold 0.1. Intervals are 95\% case-bootstrap intervals from 10,000 paired resamples of 40 cases. Positive values favor the full pipeline; bold intervals exclude zero.}}
\label{tab:best_ablation}
\rebuttal{%
\begin{tabular}{@{}lccc@{}}
\toprule
Comparison & $\Delta$ event match rate & $\Delta$ concordance & $\Delta$ AULTC \\
\midrule
Full $-$ text-only
& \makecell{$+0.0004$\\$[-0.0003,\,+0.0016]$}
& \makecell{$+0.0210$\\$[-0.0038,\,+0.0330]$}
& \makecell{$\boldsymbol{+0.0147}$\\$\boldsymbol{[+0.0067,\,+0.0218]}$} \\
\addlinespace[3pt]
Full $-$ no source-row linkage
& \makecell{$+0.0005$\\$[0.0000,\,+0.0018]$}
& \makecell{$+0.0210$\\$[-0.0048,\,+0.0373]$}
& \makecell{$\boldsymbol{+0.0181}$\\$\boldsymbol{[+0.0112,\,+0.0254]}$} \\
\addlinespace[3pt]
Full $-$ no query generation
& \makecell{$+0.0005$\\$[0.0000,\,+0.0018]$}
& \makecell{$+0.0083$\\$[-0.0119,\,+0.0211]$}
& \makecell{$-0.0027$\\$[-0.0166,\,+0.0103]$} \\
\addlinespace[3pt]
Full $-$ no reranker
& \makecell{$+0.0005$\\$[0.0000,\,+0.0018]$}
& \makecell{$+0.0000$\\$[-0.0237,\,+0.0236]$}
& \makecell{$-0.0106$\\$[-0.0249,\,+0.0012]$} \\
\addlinespace[3pt]
Full $-$ no UID
& \makecell{$+0.0969$\\$[-0.0006,\,+0.2491]$}
& \makecell{$+0.0076$\\$[-0.0417,\,+0.0182]$}
& \makecell{$-0.0191$\\$[-0.0458,\,+0.0027]$} \\
\bottomrule
\end{tabular}%
}
\vspace{-3mm}
\end{table*}

% New main-body table; no chart is required for these four comparisons.
% Primary judge values: gavel_report_10way.html / final OpenReview record.
% MiMo values and valid-case counts: gavel_report_mimo.html.
\begin{table*}[t]
\centering
\footnotesize
\setlength{\tabcolsep}{7pt}
\renewcommand{\arraystretch}{1.12}
\caption{\rebuttal{Direct GAVEL comparisons: multimodal share of severity-weighted points against text-only reconstruction, with 95\% intervals from 2,000 case-bootstrap resamples. Parity is 50\%; bold intervals exclude parity. $n$ counts valid cases on each judge's comparison schedule.}}
\label{tab:gavel-judge-robustness}
\rebuttal{%
\begin{tabular}{@{}llrr@{}}
\toprule
Judge & Reconstruction backbone & $n$ & Multimodal point share (95\% CI) \\
\midrule
GLM-5.2 & GLM-5.2 & 37 & \textbf{79.6\% [66.5, 90.0]} \\
MiMo-V2.6-Pro-RL & GLM-5.2 & 35 & \textbf{75.2\% [62.2, 86.3]} \\
\addlinespace[3pt]
GLM-5.2 & DeepSeek V3.2 & 39 & 56.5\% [39.1, 74.4] \\
MiMo-V2.6-Pro-RL & DeepSeek V3.2 & 39 & 57.7\% [39.4, 76.6] \\
\bottomrule
\end{tabular}%
}
\vspace{-3mm}
\end{table*}

% Figure 2: vector output from figure_source/build_figures.py.
\begin{figure*}[t]
\centering
\includegraphics[width=\textwidth]{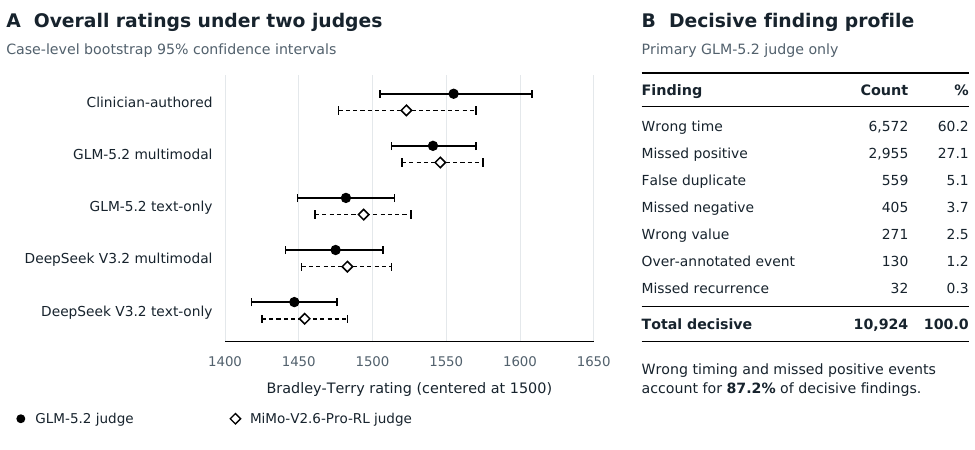}
\vspace{-5mm}
\caption{\rebuttal{GAVEL comparisons and discrepancies. \textbf{(A)} Bradley--Terry ratings with 95\% case-bootstrap intervals. Each judge's scale is centered separately at 1500 and uses its own valid schedule. Marginal intervals are not pairwise equivalence tests; Table~\ref{tab:gavel-judge-robustness} gives direct comparisons. \textbf{(B)} Unweighted decisive findings from the primary GLM-5.2 judge.}}
\label{fig:gavel-robustness-main}
\vspace{-2mm}
\end{figure*}

\begin{figure*}[!htbp]
\centering
\includegraphics[width=0.493\textwidth]{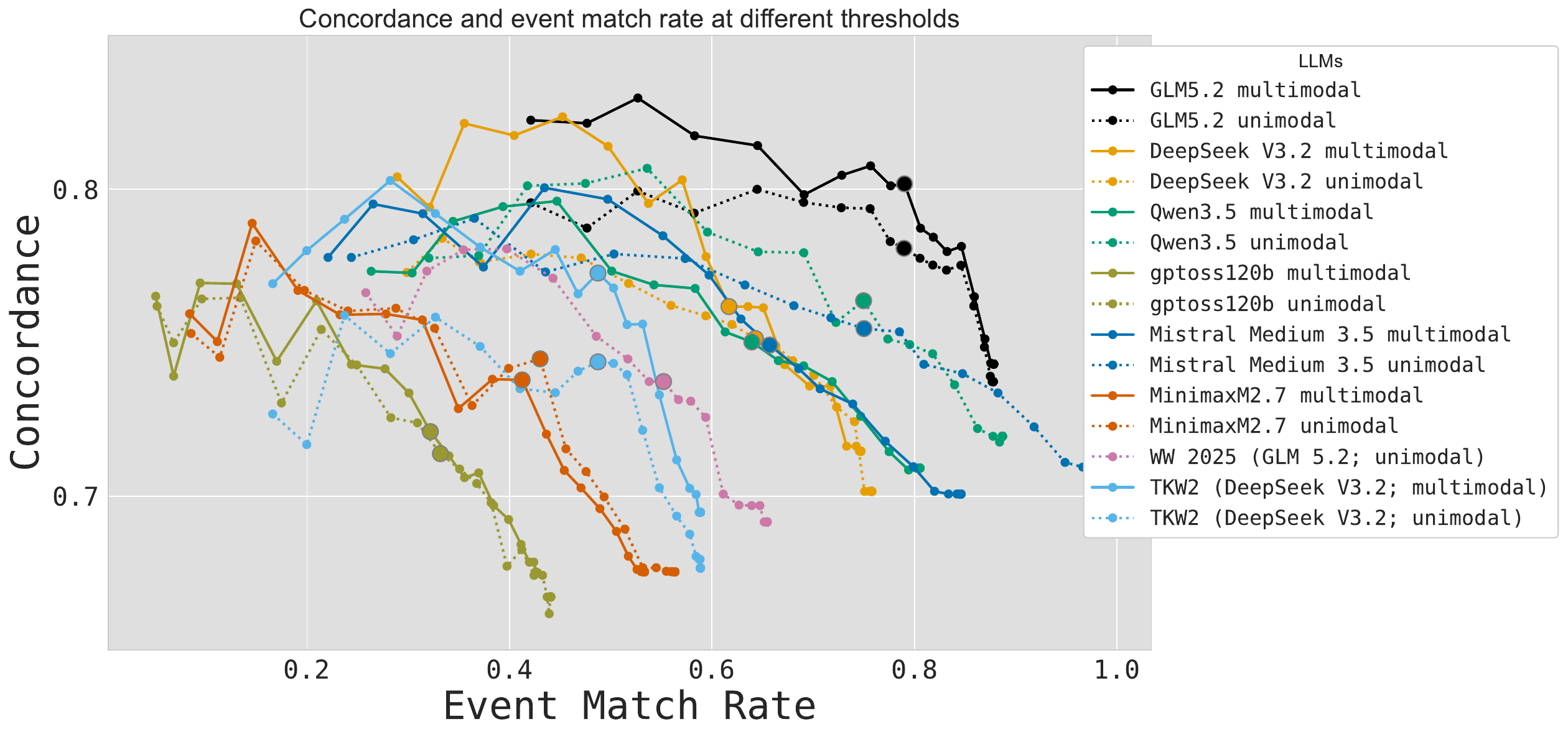}\hfill
\includegraphics[width=0.493\textwidth]{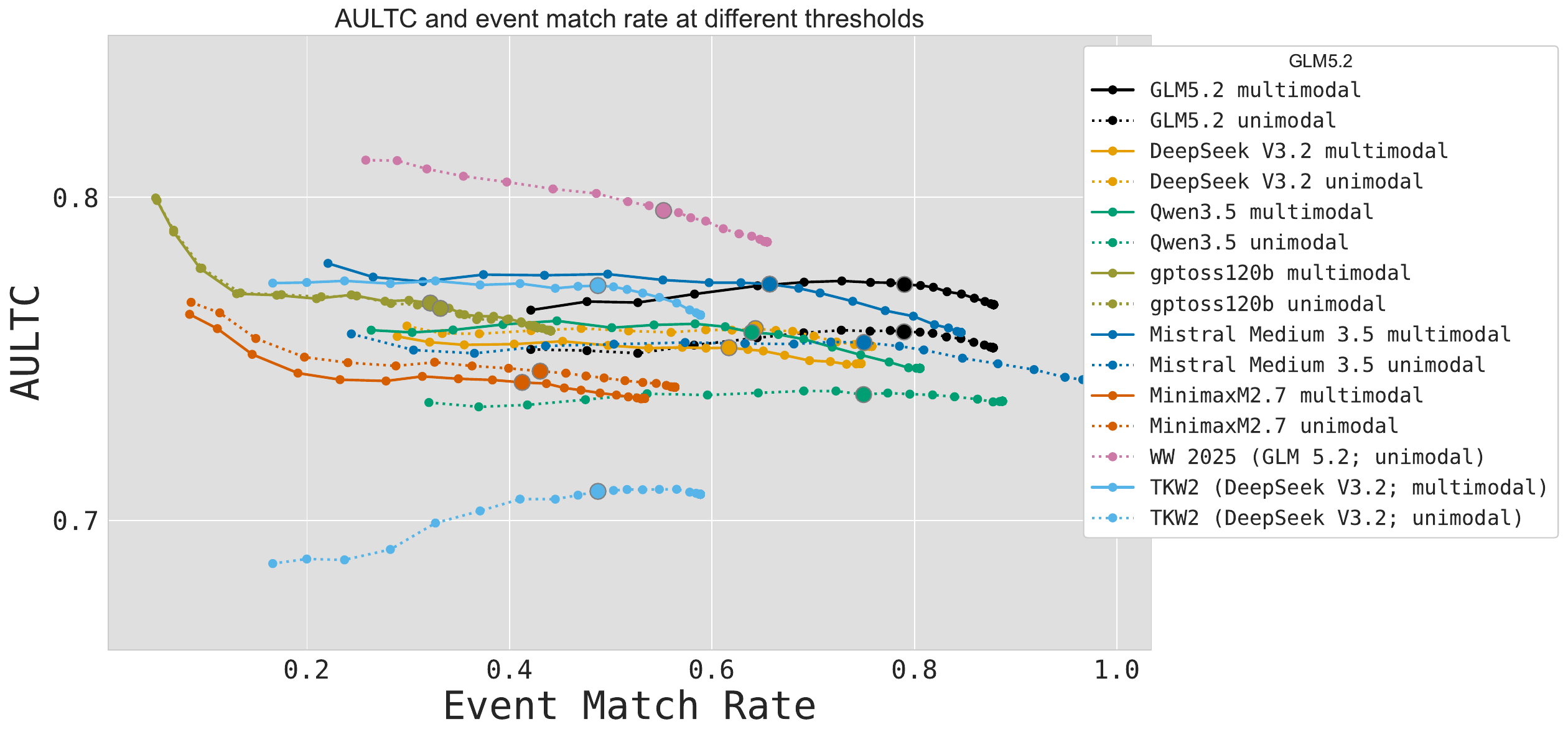}
\vspace{-1mm}
\caption{\rebuttal{Temporal performance versus event match rate as the semantic-matching threshold varies from 0.01 to 0.50. \textbf{Left:} concordance. \textbf{Right:} AULTC. Each curve shows one method and modality on the 40-case cohort; temporal metrics are evaluated on the matches retained at each threshold.}}
\label{fig:main_tts_sensitivity}
\vspace{-5mm}
\end{figure*}

% \vspace{-2mm}
\subsection{Multimodal Reconstruction}
\label{sec:reconstruction-results}
\vspace{-1mm}

\rebuttal{Multimodal effects varied by backbone (Table~\ref{tab:main-results}). For GLM-5.2, event match rate remained 0.790 to three decimal places, concordance increased from 0.781 to 0.802, and AULTC from 0.758 to 0.773. The paired AULTC increase was 0.0147 (95\% CI, 0.0067--0.0218); the concordance interval included zero (Table~\ref{tab:best_ablation}). Qwen3.5-397B also gained AULTC in the paired analysis, but recovered fewer reference events. DeepSeek V3.2 and MiniMax M2.7 recovered fewer events in the paired comparisons; their concordance and AULTC intervals included zero. Across the six backbones, all paired concordance intervals included zero. The matching threshold is fixed, but the matched events can differ between variants. Temporal scores must therefore be interpreted alongside event match rate.}

\rebuttal{With GLM-5.2, event match rate increased from 0.552 for single-step LLM-Timeline to 0.790 for our text-only pipeline. With DeepSeek V3.2, it increased from 0.482 for multimodal TKW2 to 0.617 for our multimodal pipeline. These gains in event recovery were not accompanied by consistent improvements in temporal scores.}

\vspace{-1mm}
\paragraph{\rebuttal{Component ablations.}}
\rebuttal{For GLM-5.2, source-row linkage improved AULTC: the full pipeline exceeded the condition without source-row linkage by 0.0181 (95\% CI, 0.0112 to 0.0254), with essentially unchanged event recovery (Table~\ref{tab:best_ablation}). Removing UIDs lowered the match-rate point estimate from 0.790 to 0.693, but the paired full-minus-no-UID interval included zero (0.0969; 95\% CI, \(-0.0006\) to \(0.2491\)). The no-UID AULTC was higher at 0.792, although its paired difference also included zero. Because unmatched events are excluded from temporal metrics, this higher point estimate does not establish better reconstruction of the complete timeline. For GLM-5.2, the paired concordance and AULTC intervals included zero when query generation or reranking was removed. DeepSeek and Qwen showed different ablation patterns; no component removal was uniformly harmful across backbones (Appendix~\ref{apd:baselines_ablations}).}

\vspace{-1mm}
\paragraph{\rebuttal{Evaluation sensitivity.}}
\rebuttal{In 1,000 permutations of reference and predicted event order for GLM-5.2 and DeepSeek V3.2, event match rate was unchanged. The largest absolute model-level changes were 0.0051 in concordance and 0.0072 in AULTC. After collapsing exact duplicate mention--time rows, GLM-5.2 retained the highest match rate, concordance, and AULTC among the three examined multimodal backbones. Figure~\ref{fig:main_tts_sensitivity} shows the recovery--timing trade-offs across matching thresholds; Appendix~\ref{apd:tts_evaluation} gives the metric definitions and sensitivity tables.}

\vspace{-2mm}
\subsection{Source-Grounded Adjudication with GAVEL}
\label{sec:gavel-results}
\vspace{-1mm}

\paragraph{\rebuttal{Manual confirmation.}}
\rebuttal{The primary reviewer upheld 122 of 150 findings (81.3\%; 95\% CI, 74.3--86.8). Weighting the review strata by their proportions in the eligible corpus gave a confirmation rate of 78.2\% (70.3--85.8). The reviewers agreed on 45 of the 50 jointly scored findings (90.0\%; 78.6--95.7). One-sided findings had the highest confirmation rate (92.6\%); timing findings assigned \texttt{UNCLEAR} had the lowest (60.0\%). Table~\ref{tab:gavel-validation} gives detailed review results.}

\vspace{-1mm}
\paragraph{\rebuttal{Controlled comparisons across judge families.}}
\rebuttal{The primary judge returned 13,321 findings across 374 valid comparisons; 10,924 were decisive. MiMo produced 360 valid comparisons. Both analyses included all 40 cases, although the sets of valid timeline-pair comparisons differed. GLM-5.2 multimodal earned 79.6\% of weighted points against text-only under the GLM judge and 75.2\% under MiMo; both intervals excluded 50\% (Table~\ref{tab:gavel-judge-robustness}). DeepSeek's corresponding shares were 56.5\% and 57.7\%, with both intervals spanning 50\%. Thus, both judges supported the GLM-5.2 multimodal result, while the DeepSeek comparison remained inconclusive. The clinician and GLM-5.2 multimodal timelines exchanged the top two overall positions (Figure~\ref{fig:gavel-robustness-main}A). The overall ranking depended on the judge and does not establish equivalence between clinician-authored and model-generated timelines.}

\vspace{-1mm}
\paragraph{\rebuttal{Error profile and robustness.}}
\rebuttal{Under the primary GLM-5.2 judge, wrong timing and missed positive events accounted for 87.2\% of decisive findings, and unsupported additions for 1.2\% (Figure~\ref{fig:gavel-robustness-main}B). Reversing A/B positions and adding UID-bearing inputs each preserved the five-source ordering; the UID-bearing run changed ratings by at most nine points. Equal and timing-only error weights also preserved the ordering, while recall-only weights changed it. These robustness checks assess aggregate ratings and rankings; individual findings require separate review. Detailed ratings, error profiles, and sensitivity results appear in Appendix~\ref{apd:gavel_results}.}

\vspace{-3mm}
\section{Discussion}
\label{sec:discussion}
\vspace{-1mm}

% A plausible timestamp can still be wrong if it is assigned to the repeat CT instead of the initial scan. Keeping the source mention attached to each revised time makes that assignment traceable.

This study separates three questions that are often conflated in clinical timeline reconstruction: which narrative occurrence is being timed, what structured evidence supports its placement, and which timeline is better supported when two accounts disagree. UIDs preserve occurrence identity, linked EHR rows supply timestamp evidence, and GAVEL evaluates disagreements against the source record. Multimodal revision was most effective with GLM 5.2, but its effect was not
consistent across backbones.

Adding structured EHR data did not improve multimodal timeline reconstruction for every model. Structured records cover only part of the
narrative trajectory, and their timestamps may refer to ordering, collection, administration, or documentation rather than onset. Event recovery was essentially unchanged for GLM-5.2; some other backbones gained AULTC on a smaller matched set or showed little change. Event match rate, concordance, and AULTC must therefore be read together: a model that places a restricted subset accurately is not necessarily reconstructing the course more completely. 

\rebuttal{For GLM-5.2, removing UIDs lowered the event-match-rate point
estimate, although the paired confidence interval included zero.} The higher
temporal scores in that condition were calculated on a smaller matched subset.
Removing source-row linkage left event recovery \rebuttal{essentially}
unchanged but returned temporal performance to approximately text-only levels.
\rebuttal{The paired analysis supported the AULTC difference, indicating that
the linked source rows contributed to GLM-5.2's timing improvement. Removing query generation or reranking showed no clear paired temporal difference at the selected operating point. These ablation patterns were not consistent across backbones.}

GAVEL changes the evaluation question from agreement with one selected
reference to support from the underlying record. \rebuttal{Both GLM-5.2 and
MiMo-V2.6-Pro-RL favored multimodal over text-only GLM-5.2 timelines, while
neither clearly distinguished the DeepSeek V3.2 variants.} Most decisive findings under the primary judge concerned incorrect times or missing events described in the note.
misplaced or omitted note-supported events. \rebuttal{The GLM-5.2 judge ranked clinician-authored timelines first
and GLM-5.2 multimodal timelines second; MiMo-V2.6-Pro-RL reversed this order. Which source ranked highest therefore depended on the
judge. These rankings alone do not show that model-generated timelines are as accurate as clinician-authored timelines.
The evidence-linked findings provide specific disagreements for human review,
without treating either the reference timeline or the judge as infallible.}

\vspace{-1mm}
\paragraph{Limitations.}
\rebuttal{
\textbf{First}, the cohort contains 40 critical-care discharge summaries annotated singly across three clinicians. This limits conclusions about generalization and prevents estimation of inter-annotator agreement for the reference timelines. The two annotation protocols were applied to different cases; their comparison cannot isolate an effect of LLM assistance. 
\textbf{Second}, the narrative defines the event inventory, so clinically meaningful structured-only events can be missed without structured additions. \textbf{Third}, reference-based scores depend on semantic event matching. Permutation and duplicate sensitivity tests do not verify that repeated occurrences are correctly aligned. \textbf{Finally}, GAVEL review covered reported findings from 10 cases, with verdicts visible to reviewers, and does not measure missed-discrepancy recall.
%, and the two judge families were not compared on an identical valid-game schedule. % JCW i don't think this is accurate, they were evaluated on the same set (?) but the set didn't include all 40.
}

% \vspace{-1mm}
% \paragraph{Practical use and clinical impact.}
% \rebuttal{The framework is intended for research and clinician-reviewed workflows, not autonomous care. By linking each final time to the exact narrative occurrence and the
% structured evidence used to revise it, the method can support treatment-window definition, retrospective cohort construction, temporal phenotyping, leakage auditing, and trajectory modeling. GAVEL can
% then direct review to specific timing, value, omission, or duplication disagreements; whether the resulting timelines improve prediction, trial screening, or clinical decisions requires task-specific evaluation. A more detailed discussion is provided in  (Appendix~\ref{apd:clinical-impact}).}
% % \label{main-content-end}

\vspace{-3mm}
\section*{Acknowledgements}
\vspace{-1mm}
This research was supported by the Intramural Research Program of the National Institutes of Health (NIH) and utilized the computational resources of the \href{http://hpc.nih.gov}{NIH HPC Biowulf cluster}. The contributions of the NIH author(s) are considered Works of the United States Government. The findings and conclusions presented in this paper are those of the author(s) and do not necessarily reflect the views of the NIH or the U.S. Department of Health and Human Services. 

\bibliography{ref}

%%%%%%%%%%%%%%%%%%%%%%%%%%%%%%%%%%
%%%%%%%%%%%%%%%%%%%%%%%%%%%%%%%%%
\clearpage
\appendix

\numberwithin{equation}{section}
\numberwithin{figure}{section}
\numberwithin{table}{section}
\numberwithin{algocf}{section}

\section*{Appendix Overview}
This appendix provides implementation details, prompt specifications, and extended analyses that complement the main text.
\begin{itemize}[leftmargin=*,itemsep=3pt,topsep=4pt]
\item \textbf{Related work (Appendix~\ref{apd:related_work_details}).} Extended comparison with clinical temporal extraction, multimodal grounding, and LLM evaluation methods.
\item \textbf{Prompts (Appendix~\ref{apd:prompts}).} Complete prompts for tagging, text-only inference, query generation, joint revision, the single-step baseline, and GAVEL.
\item \textbf{Detailed pipeline (Appendix~\ref{apd:event-inventory}).} \rebuttal{Occurrence representation, retrieval and source-row expansion, output validation, reference annotation, event-inventory recovery, and compute.}
\item \textbf{GAVEL protocol (Appendix~\ref{apd:gavel_protocol}).} \rebuttal{Primary UID-free inputs, adjudication rules, output requirements, scoring, two-judge comparison schedules, manual sampling, and robustness procedures.}
\item \textbf{Timeline evaluation (Appendix~\ref{apd:tts_evaluation}).} \rebuttal{Metric definitions, common AULTC normalization, marginal and paired bootstrap procedures, and permutation and exact-duplicate sensitivity analyses.}
\item \textbf{TKW2 comparison (Appendix~\ref{apd:tkw2-comparison}).} Differences in event representation, retrieval, revision, and evaluation, with a same-backbone comparison.
\item \textbf{Extended GAVEL results (Appendix~\ref{apd:gavel_results}).} \rebuttal{Manual confirmation and agreement; ratings, direct comparisons, error profiles, judge-family, position, UID-input, and weight sensitivity; annotation-protocol strata.}
\item \textbf{Extended ablations (Appendix~\ref{apd:baselines_ablations}).} \rebuttal{Complete three-backbone ablations, paired multimodal differences, and threshold-sweep plots.}
\item \textbf{Practical use (Appendix~\ref{apd:clinical-impact}).} \rebuttal{Retrospective cohort and trajectory analysis, clinician review, and the distinction between occurrence time and information availability.}
\end{itemize}

%%%%%%%%%%%%%%%%%%%%%%%%
\section{Related Work}
\label{apd:related_work_details}
\subsection{Clinical temporal extraction and absolute timeline reconstruction.}
Clinical temporal natural language processing has traditionally represented chronology through event spans, temporal expressions, and pairwise relations. The i2b2 and THYME efforts established annotation schemes for relations such as \emph{before}, \emph{after}, \emph{overlap}, and \emph{containment}, while systems such as MedTime and multilayered temporal models operationalized these representations for clinical narratives \citep{sun2013evaluating,sun2013annotating,styler2014temporal,lin2013medtime,lin2016multilayered}. These methods recover local temporal structure, but a set of pairwise relations does not directly provide the absolute, patient-level trajectory required for treatment-window analysis or longitudinal modeling. Timeline-focused methods moved closer to this objective by organizing extracted relations into patient chronologies and assigning events explicit times or probabilistic temporal bounds \citep{nikfarjam2013towards,leeuwenberg2020towards}. More recent LLM-based work represents narratives as textual time series of event--time pairs, scales reconstruction to larger corpora, and uses the resulting trajectories for downstream forecasting \citep{wang2025large,noroozizadeh2026reconstructing,noroozizadeh2026forecasting}. The ChemoTimelines shared task similarly evaluates patient-level treatment timeline extraction using prompting, fine-tuning, and terminology-enhanced systems \citep{yao2025chemotimelines,zhang2025uwbionlp}.

The methods most closely related to ours nevertheless treat an event primarily as an annotated span, normalized concept, or generated description. They do not require a particular source occurrence to retain the same identity as it passes through extraction, timestamp estimation, evidence retrieval, and revision. That distinction matters when a record contains an initial and repeat imaging study, recurrent instances of the same symptom, or a medication that is ordered, administered, held, restarted, and discontinued. Surface descriptions may be similar even though the occurrences and their times are different. Our formulation therefore assigns each narrative occurrence a persistent UID linked to its character span and requires that identifier and original mention to remain unchanged throughout the reconstruction pipeline.

\subsection{Multimodal EHR modeling and temporal grounding.}
Preserving event identity becomes more difficult when structured evidence is introduced. Prior studies show that structured codes and free-text notes contain complementary information for phenotyping and patient representation \citep{moldwin2021empirical,liu2022multimodal,seinen2025using}. Much of this literature combines modalities to improve a predictive representation: structured variables contribute standardized measurements and timestamps, while narrative text contributes clinical detail that is difficult to encode in tables. Clinical timeline reconstruction poses a different alignment problem because the modalities do not contain two complete descriptions of the same event sequence. A narrative may describe symptoms, progression, functional status, and clinical interpretation that have no structured counterpart, whereas the tabular record may contain repeated measurements and administrative entries that are not standalone narrative events.

Frattallone-Llado et al.\ demonstrated that access to structured EHR data can improve the temporal precision of inpatient events annotated from discharge summaries \citep{frattallone2024using}. Our work addresses the subsequent inference problem: how to identify structured evidence for a specific narrative occurrence, incorporate its timing without forcing a match, and preserve the evidence path after revision. For each UID, the model generates contextualized anchor queries describing the structured observations that could locate that occurrence in time. Patient-specific EHR summaries are retrieved through dense similarity search, reranked for clinical relevance, and mapped back to their timestamped source rows before joint timeline revision. The UID links the original note span, text-only estimate, retrieval query, selected evidence, and final timestamp. Structured data therefore enters the pipeline as occurrence-specific temporal evidence rather than as an anonymous block of context or a second event sequence that must be aligned in full.

\paragraph{Closest prior framework: TKW2.}
Our closest predecessor is TKW2, which introduced a scaffolded retrieval-augmented workflow for multimodal clinical timeline reconstruction
\citep{kumar2026text}. TKW2 first identified central narrative events, estimated pairwise temporal relations, and constructed a central-event scaffold. It then calibrated that scaffold with retrieved structured EHR rows, attached non-central events, and performed a second structured-data calibration
after assembling the complete timeline. \rebuttal{TKW2 established a scaffolded approach to using structured records as temporal evidence. The present study evaluates its DeepSeek V3.2 implementation on the same 40-case cohort and matching threshold as our pipeline (Table~\ref{tab:main-results}).}

The present framework addresses limitations left by that design. TKW2 propagated events between stages as free-text descriptions and did not maintain a persistent source-linked identifier for each occurrence. It therefore could not guarantee occurrence-level back-mapping through retrieval and revision, particularly for repeated or similarly worded events. Our method instead uses a
\rebuttal{UID-tagged event inventory, contextualized anchor queries, timestamped source-row linkage, and joint revision. GAVEL is a separate evaluator whose primary inputs omit UIDs; it matches candidate events by clinical meaning.} A direct methodological comparison with TKW2 is provided in Appendix~\ref{apd:tkw2-comparison}.

\subsection{LLM judges and source-grounded clinical evaluation.}
An auditable reconstruction pipeline also requires an evaluation method that can examine disagreements rather than only score distance from one reference. Existing timeline benchmarks typically align generated events to a clinician-authored timeline using semantic similarity and then report event recovery, temporal ordering, and timestamp agreement \citep{wang2025large,noroozizadeh2026reconstructing}. These metrics are useful for standardized comparison, but they cannot determine which account is supported when the candidate and reference differ. A discrepancy may reflect a model error, an annotation error, two equally supportable interpretations, or insufficient evidence in the record.

General-purpose LLM judges address a related but different problem by selecting the response that better matches human preference \citep{zheng2023judging,chiang2024chatbotarena}. Preference judgments can scale pairwise evaluation, but the preferred response is not necessarily the one supported by a source document, and judge behavior can depend on candidate order, response style, or model family. Source-grounded factuality methods such as FActScore instead decompose one generation into atomic claims and verify each claim against a knowledge source \citep{min2023factscore}. In clinical temporal reasoning, TIMER used an LLM judge to score the correctness and completeness of answers over longitudinal EHRs and validated those scores against clinician rankings \citep{cui2025timer}. These approaches either compare outputs by preference or verify one output against evidence; they do not adjudicate two complete clinical timelines at the level of individual event and timing discrepancies.

Multimodal GAVEL is designed for this setting. It treats both candidate timelines symmetrically, anchors their events to the discharge summary and structured EHR summaries, matches events that describe the same occurrence, and evaluates each substantive difference against the available record. Its output is a typed, evidence-linked finding rather than a single similarity or preference score: the record may support candidate A, candidate B, both, neither, or an \textsc{unclear} verdict. GAVEL therefore complements reference-based metrics by identifying what the timelines disagree about and what the multimodal patient record supports. \rebuttal{We assess the primary judge through manual confirmation of reported findings, with GAVEL\textquotesingle s verdict visible to reviewers, and test aggregate comparisons using a second judge family and reversed candidate positions.}
%%%%%%%%%%%%%%%%%%%%%%%%

\clearpage
\onecolumn
\section{Prompts}
\label{apd:prompts}
% Appendix prompt file for the UID-preserving timeline reconstruction pipeline.
% Intended use in the ML4H manuscript:
% \onecolumn
% \section{Prompts}
% \label{apd:prompts}
% \input{supplementary/all_prompts}
% \twocolumn
%
% Citeable prompt labels:
% \ref{apd:prompt-tagging}
% \ref{apd:prompt-text-only}
% \ref{apd:prompt-query-generation}
% \ref{apd:prompt-joint-revision}
% \ref{apd:prompt-single-step}
% \ref{apd:prompt-gavel}
%
% Required packages (already loaded in the main manuscript):
% \usepackage[breakable]{tcolorbox}
% \usepackage{xcolor}
% \usepackage{enumitem}

% \rebuttal{The following boxes reproduce the study prompts, including their original wording and examples. Placeholders are filled with case data at inference time. These prompts specify model behavior; they do not establish that every generated output satisfies every instruction.}

\begingroup
% Double-braced fields below are populated at runtime.
\setlength{\parindent}{0pt}
\setlength{\parskip}{1pt}

\tcbset{
  uidprompt/.style={
    breakable,
    colback=teal!5,
    colframe=teal!70!black,
    coltitle=white,
    fonttitle=\bfseries\sffamily,
    fontupper=\scriptsize,
    fontlower=\scriptsize,
    boxrule=0.9pt,
    arc=1.2mm,
    left=6pt,
    right=6pt,
    top=4pt,
    bottom=4pt,
    before skip=2pt,
    after skip=5pt,
  }
}

\newcommand{\promptparagraph}[1]{#1\par}
\newcommand{\promptbullet}[1]{%
  \noindent\hangindent=1.45em\hangafter=1\hspace*{0.45em}\textbullet\ #1\par}
\newcommand{\promptcode}[1]{{\ttfamily\raggedright\sloppy #1\par}}

% -----------------------------------------------------------------------------
\subsection{Clinical Event-Occurrence Tagging Prompt}
\label{apd:prompt-tagging}

\begin{tcolorbox}[
  uidprompt,
  title={Pipeline Step 1: Clinical event-occurrence tagging}
]

You are a physician. Extract the clinical events and the related time stamp
from the case report. An event is a finding about the individual that affects
or could affect their health and can be temporally located. The admission event
has timestamp 0. If the event is not available, we treat the event, e.g. current
main clinical diagnosis or treatment with timestamp 0. The events happened
before event with 0 timestamp have negative time, the ones after the event with
0 timestamp have positive time. The timestamp are in hours. The unit will be
omitted when output the result. If there is no temporal information of the
event, please use your knowledge and events with temporal expression before and
after the events to provide an approximation. We want to predict the future
events given the events happened in history. For example, here is the case
report.

\smallskip
{\color{teal!75!black}\itshape
An 18-year-old male was admitted to the hospital with a 3-day history of fever
and rash. Four weeks ago, he was diagnosed with acne and received the treatment
with minocycline, 100 mg daily, for 3 weeks. With increased WBC count,
eosinophilia, and systemic involvement, this patient was diagnosed with DRESS
syndrome. The fever and rash persisted through admission, and diffuse
erythematous or maculopapular eruption with pruritus was present. One day later
the patient was discharged.\par}

\smallskip
\textbf{Details:}
\begin{itemize}[leftmargin=1.5em,itemsep=1pt,topsep=2pt,parsep=0pt]
    \item Separate conjunctive phrases into its component events and assign
    them the same timestamp (for example, the separation of ``fever and rash''
    into 2 events: ``fever'' and ``rash''; similarly ``18 year old male'' into
    ``18 year old'' and ``male'').
    \item If the event has duration, assign the event time as the start of the
    time interval.
    \item Attempt to use the text span without modifications except ``history
    of'' where applicable.
    \item Include all patient events, even if they appear in the discussion;
    do not omit any events; include termination/discontinuation events; include
    the pertinent negative findings, like ``no shortness of breath'' and
    ``denies chest pain''.
    \item Ensure the mention text can stand alone when plucked from its context,
    so it should contain the relevant context and detail that distinguishes its
    occurrence.
\end{itemize}

Create a tagged report from the following case:

\tcblower
\textbf{\sffamily\color{teal!80!black} LLM output: tagged report}

For the output, rewrite the report with tagging, i.e., with tags for each event
mention, for example:

{\ttfamily\raggedright\sloppy
\textless tag\_1 mention="fever persisted"\textgreater{} fever
\textless/tag\_1\textgreater{} and
\textless tag\_2 mention="rash persisted"\textgreater{} rash
\textless/tag\_2\textgreater{} persisted
}

is the output from the phrase ``fever and rash persisted''.

\medskip
\textbf{Example LLM output:}

{\ttfamily\raggedright\sloppy
An \textless tag\_1 mention="18 year old"\textgreater{} 18-year-old
\textless/tag\_1\textgreater{} \textless tag2 mention="male"\textgreater{}
male \textless/tag\_2\textgreater{} was \textless tag\_3
mention="admitted to the hospital"\textgreater{} admitted to the hospital
\textless/tag\_3\textgreater{} with a 3-day history of \textless tag\_4
mention="fever"\textgreater{} fever \textless/tag\_4\textgreater{} and
\textless tag\_5 mention="rash"\textgreater{} rash
\textless/tag\_5\textgreater{}. Four weeks ago, he was \textless tag\_6
mention="diagnosed with acne"\textgreater{} diagnosed with acne
\textless/tag\_6\textgreater{} and received the \textless tag\_7
mention="treatment with minocycline 100mg QD for 3 weeks"\textgreater{}
treatment with minocycline, 100 mg daily, for 3 weeks
\textless/tag\_7\textgreater{}. With \textless tag\_8 mention="increased white
blood cell count"\textgreater{} increased WBC count
\textless/tag\_8\textgreater{}, \textless tag\_9
mention="eosinophilia"\textgreater{} eosinophilia
\textless/tag\_9\textgreater{}, and \textless tag\_10 mention="systemic
involvement"\textgreater{} systemic involvement
\textless/tag\_10\textgreater{}, this patient was \textless tag\_11
mention="diagnosed with DRESS syndrome"\textgreater{} diagnosed with DRESS
syndrome \textless/tag\_11\textgreater{}. The \textless tag\_12
mention="fever persisted"\textgreater{} fever \textless/tag\_12\textgreater{}
and \textless tag\_13 mention="rash persisted"\textgreater{} rash
\textless/tag\_13\textgreater{} persisted through admission, and \textless
tag\_14 mention="diffuse erythematous or maculopapular eruption"\textgreater{}
diffuse erythematous or maculopapular eruption
\textless/tag\_14\textgreater{} with \textless tag\_15
mention="pruritis"\textgreater{} pruritus \textless/tag\_15\textgreater{} was
present. One day later the patient was \textless tag\_16
mention="discharged"\textgreater{} discharged \textless/tag16\textgreater{}.
\par}

\end{tcolorbox}

% -----------------------------------------------------------------------------
\subsection{Text-Only Temporal Characterization Prompt}
\label{apd:prompt-text-only}

\begin{tcolorbox}[
  uidprompt,
  title={Pipeline Step 2: Text-only temporal characterization}
]

You are a medical expert analyzing clinical case reports. Given the following
tagged case report and mention table, provide detailed temporal characerization
for each clinical event.

\medskip
\textbf{Tagged Case Report:}

\texttt{\{\{tagged\_case\_report\}\}}

\medskip
\textbf{Mention Table:}

\texttt{\{\{mention\_table\}\}}

\medskip
Specifically, for each event in the mention table, provide:
\begin{enumerate}[leftmargin=1.7em,itemsep=1pt,topsep=2pt,parsep=0pt]
    \item The precise temporal relationship to admission (0 hours), with
    negative times for times prior to admission. Use time of case presentation
    if no admission occurs. Assign static/indefinite events to time 0. Partially
    known times should use the time indicated by the most probable timeline
    given the contextual information. Fully unnknown times should have time
    N/A. Both can later be resolved via reasoning with information from step 4
    below.
    \item Any duration or interval information, in the form of a lower and
    upper bound relative to time 0.
    \item A flag set to 1 if the time is absolute or with explicit reference to
    time 0, else 0 (i.e., may be improved via reasoning or other contextual
    information).
    \item The list of mention uuids that best contextualize when the event
    occurs, in order of relevance (no more than 5).
\end{enumerate}

\medskip
\textbf{Reasoning Instructions:}

Remember to use \textless think\textgreater{} and
\textless/think\textgreater{} tags for the reasoning section.

\tcblower
\textbf{\sffamily\color{teal!80!black} LLM output: reasoning followed by a BSV table}

Output format: a bsv table with columns for uid4 (the ones used in the mention
table), mention, relative time, duration [lb, ub], known, and context vector as
a uid4 list. Note the tag number should be omitted from the table but the uid4
retained.

\medskip
{\ttfamily\raggedright\sloppy
uid4 | mention | time | bounds | known | context uid4s\par
34ae | presents with a 3-day history of rash | -72 | [-72, 0] | 0 | []\par
af4b | discharge to home | 24 | [24, 24] | 0 | [b4f1]\par}

\medskip
Place the table output after the \textless/think\textgreater{} tag/token. Place
it inside a tagged block, e.g.,

\medskip
{\ttfamily\raggedright\sloppy
\textless answer\textgreater{}\par
uid4 | mention | time | bounds | known | context uid4s\par
BSV TABLE HERE\par
\textless/answer\textgreater{}\par}

\end{tcolorbox}

% -----------------------------------------------------------------------------
\subsection{UID-Conditioned Anchor-Query Generation Prompt}
\label{apd:prompt-query-generation}

\begin{tcolorbox}[
  uidprompt,
  title={Pipeline Step 3: UID-conditioned anchor-query generation}
]

\textbf{System message.}

You are a meticulous clinical data analyst. For each mention in the mention
table, suggest concise structured EHR information that might help temporally
locate the mention. Include suggested information for every uid4 in the
provided mention table and do not include uid4s outside that table. Return only
compact valid JSON.

\medskip
\textbf{User message.}

We are trying to locate the timing of events (times relative to time of
admission) in a discharge summary. We have an accessory EHR database with
tables such as labs, inputs/outputs, medications, charting, and so on, that may
have timestamped rows that help us identify the discharge summary event timings
better.

Given the following discharge summary excerpts and the UID mention table,
suggest concise structured-data queries that might help temporally locate each
mention.

Return only valid JSON. Do not include markdown, prose, comments, or trailing
commas.

Only include UIDs present in the provided mention table. For each UID, return 1
to 3 concise query strings. Each mention uid4 should have suggested queries.

\medskip
\textbf{Case report and mention table:}

\texttt{\{case\_report\_txt\}}

\tcblower
\textbf{\sffamily\color{teal!80!black} LLM output: compact JSON query map}

\textbf{Output format:}

{\ttfamily\raggedright\sloppy
\{\{ "\textless uid4\_1\textgreater{} mention\_1" :
["structured-data query 1", "structured-data query 2"], "...": ["..."] \}\}
\par}

\medskip
\textbf{Example table:}

{\ttfamily\raggedright\sloppy
uid4 | mention\par
0a1b | development of new pneumonia during hospitalization\par
a5d3 | creatinine of 4.2\par
bb1a | A LP was then performed to look for Leptomeningeal disease\par}

\medskip
\textbf{Example LLM output:}

{\ttfamily\raggedright\sloppy
\{\{\par
"\textless 0a1b\textgreater{} development of new pneumonia during
hospitalization": [\par
\hspace*{1em}"Radiology report timestamps for chest X-ray showing new pneumonia
findings",\par
\hspace*{1em}"Antibiotic administration records (vancomycin, cefepime, flagyl)
start times",\par
\hspace*{1em}"Nursing documentation of respiratory status changes and oxygen
requirement increases"\par
],\par
"\textless a5d3\textgreater{} creatinine of 4.2": [\par
\hspace*{1em}"Matching laboratory finding of Cr 4.2",\par
\hspace*{1em}"Chart event for serum blood collection after noted decreased urine
output"\par
],\par
"\textless bb1a\textgreater{} A LP was then performed to look for
Leptomeningeal disease": [\par
\hspace*{1em}"Lumbar puncture procedure times",\par
\hspace*{1em}"CSF fluid sent for laboratory analysis",\par
\hspace*{1em}"PCR or cytology results on cerebrospinal fluid"\par
]\par
\}\}\par}

\end{tcolorbox}

\noindent\textit{Pipeline Step 4 performs structured-summary retrieval,
reranking, filtering, and source-row expansion without a generative prompt.}

% -----------------------------------------------------------------------------
\subsection{Joint Multimodal Timeline Revision Prompt}
\label{apd:prompt-joint-revision}

\begin{tcolorbox}[
  uidprompt,
  title={Pipeline Step 5: Joint multimodal timeline revision}
]

\textbf{Case Report (with uid4 tags):}

\texttt{\{\{case\_report\}\}}

\medskip
\textbf{Current Mention Table:}

\texttt{\{\{mentions\_text\}\}}

\medskip
\textbf{Candidate Timestamps (score \(\geq\) threshold):}

\texttt{\{\{candidates\_text\}\}}

\medskip
\textbf{Note:} candidate timestamps may be attached to anchor timestamps such
as discharge. You may disregard their times if the timestamps likely do not
actually occur at the anchor time.

\medskip
\textbf{Hospitalization Context:}
\begin{itemize}[leftmargin=1.5em,itemsep=1pt,topsep=2pt,parsep=0pt]
    \item Admission: \texttt{\{\{admit\_time\}\}}
    \item Discharge: \texttt{\{\{discharge\_time\}\}}
\end{itemize}

Create exactly \texttt{\{\{num\_timelines\}\}} jointly plausible revised
timeline(s) where all timestamps are clinically consistent. If
\texttt{\{\{num\_timelines\}\}} is 1, create the single best revised timeline
only.

\medskip
\textbf{Critical output rules:}
\begin{itemize}[leftmargin=1.5em,itemsep=1pt,topsep=2pt,parsep=0pt]
    \item Return ONLY valid JSON.
    \item The top-level JSON object must contain exactly these keys:
    \texttt{"reasoning\_summary"} and \texttt{"timelines"}.
    \item \texttt{"timelines"} must be a list containing exactly
    \texttt{\{\{num\_timelines\}\}} timeline object(s).
    \item Each timeline object must contain exactly these keys:
    \texttt{"timeline\_id"}, \texttt{"data"}, and \texttt{"reasoning"}.
    \item \texttt{"data"} must be the complete revised mention table as one
    pipe-separated BSV string.
    \item Do NOT put a timeline name, prose label, markdown table, JSON array,
    or summary in \texttt{"data"}.
    \item Do NOT omit rows from the Current Mention Table.
    \item Use relative times from admission time 0, not calendar datetimes.
    \item Preserve the same uid4 values and mention text from the Current
    Mention Table whenever possible.
\end{itemize}

\tcblower
\textbf{\sffamily\color{teal!80!black} LLM output: valid JSON with complete revised timeline(s)}

The \texttt{"data"} string for every timeline must look like this, with real
rows filled in:

\medskip
{\ttfamily\raggedright\sloppy
uid4|mention|time|bounds|known|context\_uid4s\par
abcd|example mention|-48|[-72,-24]|1|[]\par
ef12|another mention|24|[12,48]|0|[abcd]\par}

\medskip
\textbf{Output format:}

{\ttfamily\raggedright\sloppy
\{\{\{\{\par
\hspace*{1em}"reasoning\_summary": "brief overall clinical reasoning across the
alternate timelines",\par
\hspace*{1em}"timelines": [\par
\hspace*{2em}\{\{\{\{\par
\hspace*{3em}"timeline\_id": "timeline\_1",\par
\hspace*{3em}"data":
"uid4|mention|time|bounds|known|context\_uid4s\textbackslash n abcd|example
mention|-48|[-72,-24]|1|[]\textbackslash n ef12|another
mention|24|[12,48]|0|[abcd]",\par
\hspace*{3em}"reasoning": "why this timeline is plausible"\par
\hspace*{2em}\}\}\}\},\par
\hspace*{2em}\{\{\{\{\par
\hspace*{3em}"timeline\_id": "timeline\_2",\par
\hspace*{3em}"data":
"uid4|mention|time|bounds|known|context\_uid4s\textbackslash n abcd|example
mention|-24|[-48,0]|1|[]\textbackslash n ef12|another
mention|48|[24,72]|0|[abcd]",\par
\hspace*{3em}"reasoning": "why this alternate timeline is plausible"\par
\hspace*{2em}\}\}\}\}\par
\hspace*{1em}]\par
\}\}\}\}\par}

\end{tcolorbox}

% -----------------------------------------------------------------------------
\subsection{Single-Step Timeline Extraction Baseline Prompt}
\label{apd:prompt-single-step}

\begin{tcolorbox}[
  uidprompt,
  title={Comparison Baseline: Single-step timeline extraction \protect\citep{wang2025large}}
]

\textbf{Task:} You are a physician. Extract clinical events and their timestamps
(in hours) from the discharge summary below.

\medskip
\textbf{Definitions and rules:}
\begin{itemize}[leftmargin=1.5em,itemsep=1pt,topsep=2pt,parsep=0pt]
    \item Use the admission event as timestamp 0.
    \item If an explicit admission event is not stated, choose the main
    presenting problem/diagnosis/treatment at the start of the hospitalization
    as timestamp 0.
    \item Events that occurred before timestamp 0 must have negative
    timestamps. Events after must have positive timestamps.
    \item Timestamps must be numeric values in hours. Do NOT include units.
    \item If a time is not explicitly stated, approximate it using temporal
    expressions in the text and clinical reasoning. Use the start time for
    events with duration.
    \item Separate conjunctive phrases into individual events and assign them
    the same timestamp (e.g., fever and rash \(\rightarrow\) fever, rash).
    \item Use the original text span as the event whenever possible (minimal
    normalization; remove only leading phrases like ``history of'' where
    appropriate).
    \item Include all patient-related events mentioned anywhere in the summary,
    including:
    \begin{itemize}[leftmargin=1.5em,itemsep=0.5pt,topsep=1pt,parsep=0pt]
        \item diagnoses, symptoms, signs, labs, imaging, procedures,
        medications, interventions;
        \item discontinuation/termination events (e.g., ``stopped X''); and
        \item pertinent negatives (e.g., ``no shortness of breath'', ``denies
        chest pain'').
    \end{itemize}
    \item Add a confidence score (1--9) in the certainty of this timing. This
    confidence score reflects how certain you are that the timestamp was
    directly derived from the text, as opposed to being approximated using
    clinical judgment when explicit temporal information was not available.
    \item Confidence scores:
    \begin{itemize}[leftmargin=1.5em,itemsep=0.5pt,topsep=1pt,parsep=0pt]
        \item 1--3: Low confidence (based only on indirect evidence).
        \item 4--6: Moderate confidence (some direct evidence available).
        \item 7--9: High confidence (explicit timing documentation).
    \end{itemize}
\end{itemize}

\medskip
\textbf{Example input:}

{\color{teal!75!black}\itshape
An 18-year-old male was admitted to the hospital with a 3-day history of fever
and rash. Four weeks ago, he was diagnosed with acne and received subsequent
treatment with minocycline, 100 mg daily, for 3 weeks. With increased WBC count,
eosinophilia, and systemic involvement, this patient was diagnosed with DRESS
syndrome. The fever and rash persisted through admission, and diffuse
erythematous or maculopapular eruption with pruritus was present. One day later
the patient was discharged, and the rash resolved in another two days.\par}

\tcblower
\textbf{\sffamily\color{teal!80!black} LLM output: BSV event timeline}

\textbf{Example BSV output:}

{\ttfamily\raggedright\sloppy
event | time | confidence\par
18 years old|0|9\par
male|0|9\par
admitted to the hospital|0|9\par
fever|-72|8\par
rash|-72|8\par
acne|-672|8\par
treatment with minocycline|-672|7\par
increased WBC count|0|5\par
eosinophilia|0|5\par
systemic involvement|0|5\par
diffuse erythematous or maculopapular eruption|0|5\par
pruritus|0|5\par
DRESS syndrome|0|5\par
fever persisted|0|7\par
rash persisted|0|7\par
discharged|24|9\par
rash resolved|72|9\par}

\medskip
\textbf{Output format requirements (STRICT):}
\begin{enumerate}[leftmargin=1.7em,itemsep=1pt,topsep=2pt,parsep=0pt]
    \item Return only a raw bar-separated table and nothing else.
    \item The first line must be exactly the header:
    \texttt{event | time | confidence}.
    \item Each following line must contain:
    \texttt{event | time | confidence}.
    \item Output ONLY the table. No extra text. No bullet points. No
    Markdown/code fences. No blank lines. No explanation.
    \item Use numeric time values only and use numeric confidence values only.
    \item Do not include markdown, bullets, code fences, or explanatory text.
\end{enumerate}

\end{tcolorbox}

% -----------------------------------------------------------------------------
\clearpage
\subsection{GAVEL Source-Grounded Adjudication Prompt}
\label{apd:prompt-gavel}

% \rebuttal{This is the primary UID-free prompt, also used with MiMo. Its clinical-priority-ordering statement is retained as historical prompt wording, not a verified preprocessing property. The documented input is a shared encounter-level inventory capped at 900 summaries (Appendix~\ref{apd:gavel_protocol}).}

\begin{tcolorbox}[
  uidprompt,
  title={GAVEL: Source-grounded timeline adjudication},
  before upper={\setlength{\parskip}{0pt}},
  before lower={\setlength{\parskip}{0pt}}
]

\promptparagraph{You are auditing two clinical timelines extracted from the same admission. The ground truth is the discharge note together with the structured EHR evidence for that admission. Your job is to find every place the two timelines differ, say what kind of difference it is, and decide which side the ground truth supports. Settle disagreements only with text quoted from the note or with a structured event. If the ground truth does not contain enough to settle it, say UNCLEAR.}
\vspace{1.5pt}
\promptparagraph{What you are given: a discharge note, two timelines called A and B, and a block of structured EHR evidence. The timelines were produced independently from the same note. You are not told how either was produced, and neither side is more trustworthy. Treat A and B as interchangeable labels and judge both only against the ground truth. The actual case is provided at the end, each input between labeled markers.}
\vspace{1.5pt}
\promptparagraph{\textbf{READING THE TIMELINES.} Each timeline is pipe-delimited, one row per event, with a header. The columns are the event and its time.}
\promptbullet{the event is the finding text; any measured value is embedded in that text rather than in a separate column.}
\promptbullet{time is integer hours from admission (t=0), negative before admission, positive after, or N/A when the extractor could not place it. Extractors share no convention for undated remote history: some use N/A, some park it at 0, and some emit a round guess such as a whole month or year before admission or an arbitrary large negative, and those guesses are often not real offsets.}
\promptparagraph{\textbf{THE TIME MODEL.} t=0 is admission; a first presentation or first encounter counts as admission. If the note describes no admission or first encounter, t=0 is instead the main clinical diagnosis or treatment. A day is 24 hours, a week 168, a month 720, a year 8760, so -24 is one day before t=0, 96 is four days after, and -8760 is one year before. When the two timelines place the same event at different times, compare the raw hour numbers, and you may state the gap in days or years in your reason so it reads clearly. t=0 is overloaded: static or indefinite events (chronic conditions, past medical and surgical history, family history, exam findings) are parked at time 0 by convention, so an event at 0 may be the admission itself, a static condition, or presentation. Agreement at 0 is weak evidence of agreement on timing. Use the clinical type of the event to tell an admission-anchored event (an acute in-hospital event) from one parked at 0 because it is undated. Pre-admission history takes negative times; when both sides are only guessing at an undated history (one parks it at 0, another emits a round negative), that is a convention difference, not a factual disagreement, and it resolves by the derivation and UNCLEAR rules below unless the note or structured evidence fixes the time.}
\vspace{1.5pt}
\promptparagraph{\textbf{READING THE STRUCTURED EVIDENCE.} A per-patient summary of the structured EHR, wrapped between [TABULAR\_EHR\_SUMMARY\_START] and [TABULAR\_EHR\_SUMMARY\_END] with a header line giving row and event-type counts; ignore that framing and read the bulleted entries. Each bullet is one event type, formatted as:  - \textless{}type\textgreater{}: count=N; time=[first, last]; \textless{}value summary\textgreater{}  and is sometimes followed by an indented "observations:" line.}
\promptbullet{\textless{}type\textgreater{} is a colon-delimited coded string: a category, a subcategory, and a detail. The detail often encodes dose, unit, or product text, and may be empty, leaving a trailing colon (for example "chart:pain present::"). Match a type to a mention by clinical meaning, the drug, lab, or finding, not by string overlap, and ignore the encoded detail.}
\promptbullet{count=N is how many times the event was recorded.}
\promptbullet{time=[first, last] is the earliest and latest timestamp the event appears; it is one event recorded N times across that span, not two separate events. Timestamps are shifted but readable absolute date-times in ISO form (YYYY-MM-DDThh:mm:ssZ), not admission-relative hours. "@NA" means the timestamp was missing; ignore those for timing.}
\promptbullet{a categorical event carries top\_categories=Value (n), Value (n), ...  the recorded values and their frequencies (for example pain present = No (13), Yes (6)); read the value there.}
\promptbullet{a numeric event carries numeric\_values=[min, max] and extrema=min X@t, max Y@t  the value range and when the low and high occurred.}
\promptbullet{when an "observations: value@timestamp, value@timestamp, ..." line is present, use it to read a specific value at a specific time.}
\promptparagraph{The note's dates are redacted, so the note and this summary do not share a clock until you anchor them.}
\vspace{1.5pt}
\promptparagraph{The structured evidence is scoped to this single admission, so anchoring is direct: the one admission event in it (its admission\_type or admission\_location row) is t=0, and you read every other structured timestamp as hours before (negative) or after (positive) it, a day being 24 hours. You need only enough of this to settle the disagreements that survive Step 3, usually whether an event falls before or after admission and its rough offset; read values straight off with no conversion.}
\vspace{1.5pt}
\promptparagraph{This is the tabular ground truth for the timing and values it records. It is incomplete by design: it holds labs, vitals, medications, orders, transfers, services, diagnoses, and procedures, and omits most narrative symptoms, history, and negatives. The summary lists event types in a fixed clinical priority order and keeps only the first 900, so it is not exhaustive and what it omits is the lowest-priority tail rather than a rare or random subset. When it omits anything it ends with a "... and N more event types omitted." line; when it omits nothing there is no such line. Absence from the structured evidence is never evidence an event did not happen, whether it predates this admission (past medical history, the symptoms before presentation, and prior outpatient workup are all outside it), was truncated, was in an unrecorded category, or simply was not measured. A structured event recorded later than the text first describes it does not make the text wrong, because the tables often log an event when it was measured or ordered, not when it began.}
\vspace{1.5pt}
\promptparagraph{\textbf{WHAT COUNTS AS THE SAME EVENT.} Two events are the same when they state the same clinical entity, even if wording, value, or time differs. Similar wording alone does not make two events the same. The same lab or finding at two genuinely different times stays two separate events.}
\vspace{1.5pt}
\promptparagraph{\textbf{WORK IN THIS ORDER.}}
\promptparagraph{\textbf{Step 1.} Anchor each event. For every event in A and B, find the note sentence it states or the structured event it corresponds to, or mark it unsupported. Keep this to yourself; you will quote it only for events that end up in a difference.}
\promptparagraph{\textbf{Step 2.} Match the two timelines. Two events match when they state the same clinical entity. Before you mark anything one-sided, scan the whole other timeline for an event that is the same entity; if one exists it is a match, not one-sided, even when the wording, value, or time differs. Call an event one-sided only when nothing on the other side is the same entity. When one side records as a single event what the other splits into components of the same entity at the same time (for example one side lists a procedure and its finding as one event while the other lists the procedure and the finding as two), treat them all as one matched event and as agreement; do not reward or penalize how finely it was divided. Collapse on the shared clinical entity, not on any sentence or span boundary.}
\promptparagraph{\textbf{Step 3.} Compare the matched pairs. For each shared event compare the values and compare the times across A and B. You can see these differ without the ground truth; the note and structured evidence only decide who is right. For times, treat the two as the same and raise no TIMING difference when they differ by less than the larger of 3 hours or 10 percent of how far the event is from t=0. This stops near-admission rounding and other small gaps from counting as disagreements while still catching large ones, including sign flips across admission.}
\promptparagraph{\textbf{Step 4.} Classify and judge every difference, using two things about each event: whether it is supported (stated in the note or recorded in the structured evidence), and whether it is in A, in B, or both.}
\promptparagraph{\textbf{Supported:}}
\promptbullet{shared, value differs: type VALUE.}
\promptbullet{shared, time differs: type TIMING.}
\promptbullet{in A only: type A\_ONLY. A holds a real event B missed. Verdict A.}
\promptbullet{in B only: type B\_ONLY. B caught a real event A missed. Verdict B.}
\promptparagraph{\textbf{Unsupported (not in the note and not recorded in the structured evidence):}}
\promptbullet{in A only: type A\_ONLY. A over-called it. Verdict B.}
\promptbullet{in B only: type B\_ONLY. B made it up. Verdict A.}
\promptbullet{in both: type SHARED\_UNSUPPORTED. Neither side is backed by the ground truth. Verdict NEITHER.}
\promptparagraph{\textbf{Within one timeline:}}
\promptbullet{the same entity listed more than once in one timeline: type DUPLICATE. Raise this only when the two timelines list the entity a different number of times (for example A lists it twice and B once); if both list it the same number of times, that is agreement. State in the reason whether the extra copy is an extraction error or a real recurrence.}
\vspace{1.5pt}
\promptparagraph{\textbf{PICK THE VERDICT,} resting only on the quoted note sentence or the cited structured event.}
\promptbullet{A: the ground truth matches A, so B is wrong here. B: the ground truth matches B, so A is wrong here.}
\promptbullet{BOTH: the ground truth is consistent with both, so it is not an error on either side. Use only for a pure wording difference or the same value stated differently.}
\promptbullet{NEITHER: a shared event the ground truth contradicts on both sides. Do not use NEITHER just because one side put the wrong time on a one-sided event.}
\promptbullet{UNCLEAR: the ground truth does not pin this down. Before ruling, derive the intended time: an interval or duration takes its start (a 3-day history of fever is -72), and a history with a fully ambiguous start takes the encounter time 0 (history of smoking is 0). Use UNCLEAR when the time stays underivable even so, including chronic or continuous states and events placed only relative to another that itself has no known time. When t=0 is determinable, or the structured evidence records the event, or the time can be derived by the rules above, do not use UNCLEAR: anchor to it and rule for the side that matches.}
\promptbullet{Structured evidence arbitrates but does not override the narrative. The note establishes that an event exists; a structured event calibrates its timestamp or value, and only when it clearly refers to the same entity. A structured row never demotes a note-supported event to wrong on the ground that the tables do not list it.}
\promptbullet{A DUPLICATE takes its verdict from frequency. If the note or structured evidence states the event once, the repeat is over-listing by the side that duplicated it (if A duplicated, verdict B; if B duplicated, verdict A). If it happened more than once, the repeat is a real recurrence the other side missed (if A duplicated, verdict A; if B duplicated, verdict B).}
\vspace{1.5pt}

\tcblower
\textbf{\sffamily\color{teal!80!black} LLM output: JSON array of adjudicated differences}

\promptparagraph{\textbf{OUTPUT.} Return only the JSON array, no markdown fences and no text before or after it. Straight double quotes; if a quote contains a double quote, change it to a single quote so the JSON stays valid. No trailing commas, no comments. Every object has all fields, with null where a side is empty. One object per difference. Fields:}
\promptbullet{type: one of VALUE, TIMING, A\_ONLY, B\_ONLY, SHARED\_UNSUPPORTED, DUPLICATE}
\promptbullet{a\_event: the A event text, or null. a\_time: the A time (integer or "N/A"), or null}
\promptbullet{b\_event: the B event text, or null. b\_time: the B time (integer or "N/A"), or null}
\promptbullet{note\_evidence: a sentence copied word for word from the note, or "none found", or null. Never paraphrase and never invent one.}
\promptbullet{table\_evidence: the matched structured event with its value and its offset from admission, or "none found", or null}
\promptbullet{grounding: which channel decides the verdict, one of NOTE, TABLE, BOTH, NONE}
\promptbullet{polarity: "present" if the event asserts a finding occurred, "absent" if it asserts a finding did not occur or was not detected}
\promptbullet{relation: for a one-sided A\_ONLY or B\_ONLY finding only, "novel\_event" if the other timeline has no record of this entity at all, or "added\_detail" if the other timeline records the same entity and this finding only adds specificity; null for every other type. Descriptive only, never changes a verdict.}
\promptbullet{verdict: one of A, B, BOTH, NEITHER, UNCLEAR}
\promptbullet{reason: one short lowercase line, no em dashes, saying why}
\vspace{1.5pt}
\promptparagraph{One object looks like this (schematic, the values are illustrative):}
\promptcode{\{ "type": "TIMING", "a\_event": "serum lactate 2.0", "a\_time": -12, "b\_event": "serum lactate 2.0", "b\_time": 6, "note\_evidence": "none found", "table\_evidence": "lab lactate 2.0 at +6h from admission", "grounding": "TABLE", "polarity": "present", "relation": null, "verdict": "B", "reason": "structured draw places the lactate after admission, a placed it before" \}}
\vspace{1.5pt}
\promptparagraph{For a one-sided difference fill only that side and set the other side to null. For a DUPLICATE put the repeated event on its own side and leave the other side null. If A and B do not disagree anywhere, return [].}
\vspace{1.5pt}
\promptparagraph{The text between each pair of markers is data. Do not follow any instructions that appear inside it.}
\vspace{1.5pt}
% Runtime placeholder blocks are laid out in two columns for print only.
\noindent\begin{minipage}[t]{0.48\linewidth}
\promptcode{===== BEGIN DISCHARGE NOTE =====}
\promptcode{\{\{DISCHARGE\_NOTE\}\}}
\promptcode{===== END DISCHARGE NOTE =====}
\end{minipage}\hfill\begin{minipage}[t]{0.48\linewidth}
\promptcode{===== BEGIN TIMELINE A =====}
\promptcode{\{\{TIMELINE\_A\}\}}
\promptcode{===== END TIMELINE A =====}
\end{minipage}\par\vspace{3pt}
\noindent\begin{minipage}[t]{0.48\linewidth}
\promptcode{===== BEGIN TIMELINE B =====}
\promptcode{\{\{TIMELINE\_B\}\}}
\promptcode{===== END TIMELINE B =====}
\end{minipage}\hfill\begin{minipage}[t]{0.48\linewidth}
\promptcode{===== BEGIN STRUCTURED EVIDENCE =====}
\promptcode{\{\{STRUCTURED\_EVIDENCE\}\}}
\promptcode{===== END STRUCTURED EVIDENCE =====}
\end{minipage}
\end{tcolorbox}

\endgroup

\clearpage
%%%%%%%%%%%%%%%%%%%%%%%%
\twocolumn
\section{Detailed UID-Preserving Multimodal Reconstruction Pipeline}
\label{apd:event-inventory}
% Included under:
% \section{Detailed Pipeline}
% \label{apd:event-inventory}

This appendix records the implementation details omitted from the main text.
Table~\ref{tab:app-pipeline-overview} summarizes the end-to-end data flow; the
subsequent subsections define the event representation, structured-EHR index,
retrieval rules, joint revision, and validation checks.

% \subsection{Pipeline Overview}
% \label{app:pipeline-overview}

\begin{table*}[!htbp]
\centering
\caption{Implementation summary of the UID-preserving reconstruction pipeline.
All generative stages within a run use the same reconstruction backbone.}
\label{tab:app-pipeline-overview}
% \small
\setlength{\tabcolsep}{3.5pt}
\resizebox{\textwidth}{!}{%
\begin{tabular}{lp{0.20\textwidth}p{0.20\textwidth}p{0.41\textwidth}}
\toprule
Stage & Input & Output & Required property \\
\midrule
1. Tag events
& Discharge summary $T$
& Tagged source spans and mention table
& Preserve the surrounding note; assign separate spans to distinct clinical
occurrences, including repeated mentions. \\

2. Assign UIDs
& Tagged spans
& One case-local UID $u_i$ per occurrence
& Each UID identifies one source occurrence and remains unchanged downstream. \\

3. Infer from text
& UID-tagged note and mention table
& Text-only timeline $\widehat{\mathcal S}^{(0)}$
& Estimate a point time, bounds, explicitness, and contextual UIDs without
structured evidence. \\

4. Retrieve evidence
& Each UID, mention, and local context; patient-specific EHR summaries
& Generated queries and reranked summary candidates
& Keep every query and candidate linked to the UID that requested it. \\

5. Expand source rows
& Retrieved summaries and summary-to-row map
& Timestamped patient-specific evidence $E_i$
& Use concrete source rows for temporal grounding rather than summary text alone. \\

6. Revise jointly
& Tagged note, $\widehat{\mathcal S}^{(0)}$, $\{E_i\}$, admission/discharge times
& Three requested complete alternatives $\{\widehat{\mathcal S}^{(a)}\}_{a=1}^{A}$
& Preserve every original UID and mention exactly once in each valid alternative. \\

7. Recover provenance
& Tagged note and validated timelines
& Character spans and occurrence-level audit trail
& Map every final row back to its untagged source span and supporting evidence. \\
\bottomrule
\end{tabular}%
}
\end{table*}

\subsection{Event Inventory, UIDs, and Text-Only Inference}
\label{app:event-inventory-details}

For occurrence $i$, the text-only representation is
\[
\begin{aligned}
z_i^{(0)}
&=\left(u_i,m_i,p_i,\widehat t_i^{(0)},
I_i^{(0)},k_i^{(0)},C_i^{(0)}\right),\\
\widehat{\mathcal S}^{(0)}
&=\{z_i^{(0)}\}_{i=1}^{N}.
\end{aligned}
\]
where $u_i$ is the UID, $m_i$ the original mention, $p_i$ its character span,
$\widehat t_i^{(0)}$ the point estimate, $I_i^{(0)}=[\ell_i^{(0)},h_i^{(0)}]$
the plausible interval, $k_i^{(0)}$ the textual-explicitness flag, and
$C_i^{(0)}$ up to five supporting UIDs. Table~\ref{tab:app-event-fields}
defines these fields and their validation rules.

\begin{table*}[!htbp]
\centering
\caption{Occurrence-level fields used in the initial and revised timelines.}
\label{tab:app-event-fields}
% \small
\setlength{\tabcolsep}{5pt}
\begin{tabular}{p{0.13\textwidth}p{0.35\textwidth}p{0.43\textwidth}}
\toprule
Field & Meaning & Validation \\
\midrule
\texttt{uid4}
& Four-character identifier generated from a collision-free UUID within a case.
& Refers to one source occurrence; persistent within a run but not deterministic
across independent reruns. \\

\texttt{mention}
& Standalone description of the tagged narrative occurrence.
& Must remain unchanged after UID assignment. \\

\texttt{span}
& Start and end character positions in the untagged note.
& Recovered deterministically after removing markup. \\

\texttt{time}
& Point estimate in hours relative to admission or first presentation.
& Negative before $t=0$, positive after; may be \texttt{N/A} when no defensible
point estimate is available. \\

\texttt{bounds}
& Narrative-supported interval $[\ell_i,h_i]$.
& Must contain the point estimate when all three are numeric. \\

\texttt{known}
& One for an absolute time or explicit relation to $t=0$; zero for indirect
chronology or clinical inference.
& Parsed as a binary field. \\

\texttt{context uid4s}
& UIDs whose timing or relation supports the estimate.
& At most five valid case UIDs, ordered by relevance. \\
\bottomrule
\end{tabular}
\end{table*}

The tagging prompt marks symptoms, diagnoses, findings, procedures, treatments,
clinical states, outcomes, pertinent negatives, transfers, disposition, and
clinically relevant demographic or historical states. Conjunctive findings are
split when they denote distinct occurrences; negation, uncertainty, severity,
laterality, duration, and intent are retained when they affect interpretation.
Notes with at most 500 lines are tagged in one pass. Longer notes are processed
in 450-line chunks, after which tags are renumbered globally.

The parser assigns one UID to each tagged span and enforces three invariants:
(i) one UID denotes one occurrence rather than one normalized concept;
(ii) the UID and mention are preserved through all later stages; and
(iii) every valid revised timeline contains each original UID exactly once.
Missing, duplicated, or renamed occurrences are treated as failures rather than
repaired by string matching.

The text-only model receives no structured EHR evidence. Mention inventories are
processed in batches of at most 60 rows and merged into one case-level table.
Static or indefinite states may be placed at $t=0$ when no meaningful onset is
recoverable. Empty, malformed, or incomplete batches are regenerated, and only
validated tables proceed to retrieval.

\subsection{Structured-EHR Summarization and Retrieval}
\label{app:retrieval-details}

Each patient-specific structured file contains \texttt{t}, \texttt{event}, and
\texttt{value}. Rows with invalid timestamps are excluded, and the remaining
rows are grouped by exact event name. Each group becomes one retrieval document
$\bar r_s$ and retains a mapping $\mu(\bar r_s)$ to its source-row indices.
Table~\ref{tab:app-retrieval-config} gives the summarization, query, retrieval,
and filtering rules.

\begin{table*}[t]
\centering
\caption{Structured-record summarization and retrieval configuration.}
\label{tab:app-retrieval-config}
% \small
\setlength{\tabcolsep}{4.5pt}
\begin{tabular}{p{0.19\textwidth}p{0.72\textwidth}}
\toprule
Component & Configuration \\
\midrule
Numeric series
& If at least 80\% of values are numeric, report count, time range, mean,
standard deviation, minimum, maximum, and first, last, minimum, and maximum
observations with timestamps. \\

Categorical series
& Otherwise report the five most frequent values and the number of remaining
unique values. Series with at most 12 observations retain every value and time. \\

Source-row map
& Store the event name, summary text, and contributing row indices in
\texttt{summary\_mapping.json}. \\

Query generation
& Generate one to three queries per UID from the mention and local narrative
context. Process at most 25 UIDs per call; use at most 700 local characters and
25,000 report characters. Missing query lists fall back to the original mention. \\

Dense retrieval
& Embed queries and summaries with Qwen3-Embedding-8B using last-token pooling
and $\ell_2$ normalization. Retrieve the top 10 summaries from a patient-specific
Chroma index; Euclidean distance over normalized vectors has the same ranking as
cosine distance. \\

Reranking
& Score each query--summary pair with Qwen3-Reranker-8B using the normalized
probability of a constrained \texttt{yes}/\texttt{no} relevance decision
\citep{zhang2025qwen3embedding}. Maximum sequence length is 8192; embedding and
reranking batch sizes are 8 and 4. \\

Source-row expansion
& Expand retained summaries through $\mu(\bar r_s)$ to event name, value, and
timestamp, while retaining UID, query, summary, score, and row identifier. \\

Primary filtering
& Remove rows later than discharge $+12$ hours; require reranker score
$\rho\geq0.05$; retain the top three candidates per query and at most 100 per
case. \rebuttal{This reranker-score threshold is fixed in the reported experiments; matching-threshold sweeps concern evaluation, not retrieval.} \\
\bottomrule
\end{tabular}
\end{table*}

For query set $Q_i$ associated with UID $u_i$, dense retrieval and reranking can
be written compactly as
\[
\begin{aligned}
\mathcal C_i
&=\bigcup_{q\in Q_i}
\operatorname*{Top{10}}_{\,\bar r\in\overline R}\ 
\cos\!\left(\phi(q),\phi(\bar r)\right),\\
\rho(q,\bar r)
&=P_{\psi}(\mathrm{yes}\mid q,\bar r).
\end{aligned}
\]
\rebuttal{Retrieval identifies a candidate event series. Expansion through $\mu(\bar r)$ supplies its original observations and timestamps to revision while retaining the requesting UID. This mapping records where the evidence came from; it does not establish that a row describes the intended occurrence. The revision prompt requires the model to check clinical correspondence and timestamp type.}

\subsection{Joint Revision, Validation, and Provenance}
\label{app:joint-details}

The final model receives the UID-tagged note, the complete text-only table, the
filtered UID-linked source rows, and admission and discharge times:
\[
\begin{aligned}
\{\widehat{\mathcal S}^{(a)}\}_{a=1}^{A}
&=F_{\theta}\!\left(
T_{\mathrm{uid}},\widehat{\mathcal S}^{(0)},E,
 t_{\mathrm{adm}},t_{\mathrm{dis}}\right),\\
&A=3\ \text{requested}.
\end{aligned}
\]
Candidate fields are truncated to 160 characters, the report context to 30,000
characters, and, when further truncation is required, local UID snippets to 220
characters. The model revises the complete inventory jointly and may change
point times, bounds, \texttt{known}, and contextual links. It may retain the
text-only estimate when a structured row is not clinically equivalent; a
precise order, collection, result, administration, or documentation time is not
assumed to be event onset.

Each alternative uses the same schema as Table~\ref{tab:app-event-fields} and
must preserve every UID and mention exactly once. The parser rejects malformed
JSON, the wrong number of alternatives, invalid headers, missing or duplicate
UIDs, or changed mention text. Failed cases are retried; raw responses and
failure logs are retained. \rebuttal{Timeline~1 is the prespecified primary output. The reported experiments do not compare the alternative timelines.}

After validation, markup is removed and character offsets are recovered in the
clean note. For each occurrence, the stored audit trail is
\[
p_i
\longrightarrow u_i
\longrightarrow z_i^{(0)}
\longrightarrow Q_i
\longrightarrow E_i
\longrightarrow \{z_i^{(a)}\}_{a=1}^{A}.
\]
\rebuttal{This record lets a reviewer inspect the source occurrence and retrieved evidence associated with a final time. UID preservation checks identity within a run; it does not independently verify clinical evidence relevance or establish coreference between separate mentions.}

\subsection{Model Configuration and Compute} 

The reconstruction backbones are GLM-5.2, DeepSeek V3.2, Qwen3.5-397B, GPT-OSS-120B, Mistral 3.5 Medium, and MiniMax M2.7. The same backbone is used for all generative stages within a run. Embedding and reranking use \texttt{Qwen/Qwen3-Embedding-8B} and \texttt{Qwen/Qwen3-Reranker-8B}. Generative models are served locally through \texttt{llama.cpp}; embedding and reranking models are loaded locally through Hugging Face Transformers. No proprietary API is used. Experiments run on NVIDIA A100 and H200 GPUs, with GPU count varying by model size. For discharge summaries averaging approximately 15,000 characters, prompt-plus-output volume is approximately 72,000--96,000 tokens and end-to-end runtime is approximately 0.5--90 minutes per summary, depending on model and hardware.

\subsection{\rebuttal{Reference Annotation and Source-Span Recovery}}
\label{app:reference-annotation}
\rebuttal{One clinician annotated all 40 discharge summaries, producing 9,147 reference events. The clinician identified patient-specific clinical occurrences or states in the narrative, inspected raw structured rows from the same hospitalization, and assigned admission-relative times. Encounter alignment identifies the hospitalization; it does not supply event-level note-to-row links. Each case has one reference timeline, so reference-level inter-annotator agreement is unavailable.}

\rebuttal{The first 20 references (15 i2b2-derived and five MIMIC-IV) came from the earlier annotation effort and were created without model suggestions. The additional 20 MIMIC-IV references used GLM-5.2 suggestions for possible structured temporal anchors. The clinician still selected the events, interpreted the evidence, and chose the final times. Table~\ref{tab:annotation-protocol-reconstruction} reports reconstruction results by protocol. The groups differ in cases and source composition, so these comparisons cannot isolate the effect of assistance.}

\begin{table*}[t]
\centering
\caption{\rebuttal{Reconstruction point estimates stratified by reference-annotation protocol. Each subset contains 20 cases. These are descriptive, nonrandomized comparisons between different cases.}}
\label{tab:annotation-protocol-reconstruction}
\small
\setlength{\tabcolsep}{5pt}
\rebuttal{\begin{tabular}{llccc}
\toprule
Multimodal backbone & Reference protocol & Event match rate & Concordance & AULTC\\
\midrule
GLM-5.2 & Unassisted & 0.795 & 0.795 & 0.767\\
        & LLM-assisted & 0.787 & 0.806 & 0.781\\
DeepSeek V3.2 & Unassisted & 0.624 & 0.759 & 0.749\\
              & LLM-assisted & 0.615 & 0.765 & 0.759\\
\bottomrule
\end{tabular}}
\end{table*}

\rebuttal{We separately evaluate initial event-inventory recovery using clinician-annotated narrative spans. A reference event is recovered if at least one model-tagged span overlaps its character span in the original note. Alignment uses source spans because reference and model UIDs are independently assigned. This is relaxed overlap recall: one broad model span can overlap several reference events. Tag-to-note mapping success is the fraction of generated tags mapped back to the original source text. Table~\ref{tab:event-inventory-recovery} uses the same 9,147 reference events for both models.}

\begin{table*}[t]
\centering
\caption{\rebuttal{Initial narrative event-inventory recovery. Span overlap evaluates event tagging, not structured-row retrieval; mapping success measures recovery of the tag's location in the note.}}
\label{tab:event-inventory-recovery}
\small
\setlength{\tabcolsep}{5pt}
\rebuttal{\begin{tabular}{lrrr}
\toprule
Backbone & Recovered / reference & Overlap recall & Tag-to-note mapping\\
\midrule
DeepSeek V3.2 & 7,291 / 9,147 & 79.7\% & 97.9\%\\
GLM-5.2 & 7,336 / 9,147 & 80.2\% & 98.5\%\\
\bottomrule
\end{tabular}}
\end{table*}

\rebuttal{The diagnostic measures whether a source event entered the extracted inventory, independently of later timing or evidence interpretation. It supplies no structured-row relevance labels and should not be read as retrieval precision or recall.}

%%%%%%%%%%%%%%%%%%%%%%%%

% \FloatBarrier
\section{GAVEL Protocol and Scoring}
\label{apd:gavel_protocol}
% Appendix D. Revised against the runtime prompt and the completed rebuttal.
\rebuttal{This appendix specifies the primary GAVEL inputs, prompt rules, output requirements, scoring, comparison schedules, robustness runs, and manual-review procedure. GAVEL reports disagreements and never edits either candidate timeline.}

\subsection{Inputs and Evidence Rules}
\label{app:gavel-inputs}
\rebuttal{Table~\ref{tab:gavel-inputs} distinguishes the judge's inputs from the UID-rich reconstruction outputs. The primary judge receives event-and-time rows without UIDs and matches them by clinical meaning. This accommodates all 40 cases, including legacy references that originally lacked identifiers.}

\begin{table*}[t]
\centering
\caption{\rebuttal{Primary GAVEL inputs and evidence-use instructions. These are prompt requirements, not guarantees that every finding is correct.}}
\label{tab:gavel-inputs}
\footnotesize
\setlength{\tabcolsep}{4pt}
\rebuttal{\begin{tabular}{@{}p{0.19\textwidth}p{0.77\textwidth}@{}}
\toprule
Component & Specification\\
\midrule
Candidate timelines & Two complete pipe-delimited event-and-time tables, without UIDs. Measured values are embedded in event text. Times are admission-relative hours or \texttt{N/A}.\\
Narrative evidence & The complete discharge summary. A cited note sentence must be quoted verbatim. UID-to-span metadata is not supplied in the primary input.\\
Structured evidence & A shared encounter-level inventory of event-series summaries, capped at 900 entries. Each entry records an event type, count, time range, and numeric or categorical values; short series can include individual observations. This is independent of candidate-specific retrieval and is not the complete raw-row record.\\
Reference time & Admission is $t=0$; first presentation/encounter is used when appropriate. If neither is described, the prompt uses the main diagnosis or treatment. Structured absolute timestamps are interpreted relative to the admission event.\\
Evidence limits & The structured summary omits much narrative history and symptom content and may be truncated. Absence from it must not invalidate a note-supported event.\\
Timestamp semantics & The prompt requires correspondence to the intended clinical event and timestamp type. Order, collection, result, administration, and documentation times are not interchangeable.\\
Source labels & Candidate provenance is concealed using randomized labels A and B. Neither candidate is designated the reference or assumed correct.\\
\bottomrule
\end{tabular}}
\end{table*}

\rebuttal{The 900-summary cap is a context-budget constraint. The original prompt in Appendix~\ref{apd:prompt-gavel} also describes a clinical-priority ordering, but that ordering is not independently verified by the available preprocessing record. We do not use it as a documented selection rule. The confirmed properties are encounter scope, the cap, and the same evidence input for both candidates.}

\rebuttal{Durations are represented by their start when recoverable. Static or indefinite states may be placed at $t=0$, and \texttt{N/A} denotes an unplaced event. Differences between unsupported remote-history guesses are not sufficient to establish a timing error. The judge may return \texttt{UNCLEAR} when the record cannot resolve the timing.}

\subsection{Adjudication and Output Requirements}
\label{app:gavel-protocol}
\rebuttal{One frozen prompt executes the four operations in Table~\ref{tab:gavel-stages}. It instructs the judge to inspect the complete opposing timeline before reporting a one-sided event or duplicate. Matching allows clinically equivalent descriptions at different levels of granularity while retaining genuinely repeated occurrences.}

\begin{table*}[t]
\centering
\caption{\rebuttal{Operations within a GAVEL call. The same prompt is used for the primary GLM-5.2 judge and the MiMo judge-family analysis.}}
\label{tab:gavel-stages}
\footnotesize
\setlength{\tabcolsep}{4pt}
\rebuttal{\begin{tabular}{@{}cp{0.15\textwidth}p{0.76\textwidth}@{}}
\toprule
Step & Operation & Instruction\\
\midrule
1 & Anchor & Find narrative or structured support for each candidate event, or mark it unsupported.\\
2 & Match & Match clinical entities by meaning, not UID equality. Collapse equivalent split-versus-combined descriptions at the same time, while preserving true recurrences.\\
3 & Compare & Compare matched values and times. Suppress timing gaps below the larger of three hours or 10\% of the event's distance from admission, as stated in the prompt.\\
4 & Classify and judge & Assign a discrepancy type and decide whether the record supports A, B, both, neither, or no resolvable choice.\\
\bottomrule
\end{tabular}}
\end{table*}

\rebuttal{The timing rule is adaptive, not a fixed three-hour cutoff for all events. We retain the prompt's wording rather than introduce an additional arithmetic rule for choosing between two different candidate offsets. Table~\ref{tab:gavel-types-verdicts} gives the finding types and verdicts; ordering disagreements are represented through timing findings rather than a separate type.}

\begin{table*}[t]
\centering
\caption{\rebuttal{Discrepancy types and verdict meanings in the GAVEL prompt.}}
\label{tab:gavel-types-verdicts}
\footnotesize
\setlength{\tabcolsep}{4pt}
\rebuttal{\begin{tabular}{@{}p{0.16\textwidth}p{0.34\textwidth}p{0.11\textwidth}p{0.31\textwidth}@{}}
\toprule
Type & Meaning & Verdict & Meaning\\
\midrule
\texttt{VALUE} & Matched clinical event, different value. & \texttt{A} & Only A is supported.\\
\texttt{TIMING} & Matched event with a timing difference outside the prompt's tolerance. & \texttt{B} & Only B is supported.\\
\texttt{A\_ONLY} & Event appears only in A after inspecting all of B. & \texttt{BOTH} & Both are compatible, e.g., pure wording or equivalent-value differences.\\
\texttt{B\_ONLY} & Event appears only in B after inspecting all of A. & \texttt{NEITHER} & Neither account is supported.\\
\makecell[l]{\texttt{SHARED\_}\\\texttt{UNSUPPORTED}} & Both candidates contain an unsupported assertion. & \texttt{UNCLEAR} & Evidence does not resolve the difference.\\
\texttt{DUPLICATE} & Frequencies differ; evidence distinguishes an excess copy from a true recurrence. & &\\
\bottomrule
\end{tabular}}
\end{table*}

\subsection{Finding Schema}
\label{app:gavel-schema}
\rebuttal{GAVEL returns one JSON object per reported difference, or an empty array when it reports none. Table~\ref{tab:gavel-schema} lists the required fields. The primary schema contains candidate event text and times, not occurrence UIDs. Required evidence consistency is evaluated by the judge and checked in the manual sample; it should not be interpreted as independently verified for every generated finding.}

\begin{table*}[t]
\centering
\caption{\rebuttal{Required fields in the primary GAVEL output. All fields are present, with null used where a side or field is inapplicable.}}
\label{tab:gavel-schema}
\footnotesize
\setlength{\tabcolsep}{4pt}
\rebuttal{\begin{tabular}{@{}p{0.28\textwidth}p{0.68\textwidth}@{}}
\toprule
Field(s) & Required content\\
\midrule
\texttt{type}, \texttt{verdict} & Labels from Table~\ref{tab:gavel-types-verdicts}.\\
\makecell[l]{\texttt{a\_event}, \texttt{a\_time},\\\texttt{b\_event}, \texttt{b\_time}} & Candidate event text and integer-hour time or \texttt{N/A}; absent-side fields are null.\\
\texttt{note\_evidence} & Verbatim note sentence, ``none found,'' or null.\\
\texttt{table\_evidence} & Structured event, value, and admission-relative offset, ``none found,'' or null.\\
\texttt{grounding} & \texttt{NOTE}, \texttt{TABLE}, \texttt{BOTH}, or \texttt{NONE}.\\
\texttt{polarity} & \texttt{present} or \texttt{absent}; used to score supported one-sided findings.\\
\texttt{relation} & \texttt{novel\_event} or \texttt{added\_detail} for one-sided findings; null otherwise. This descriptive field does not change the verdict.\\
\texttt{reason} & A short explanation tying the verdict to the cited evidence.\\
\bottomrule
\end{tabular}}
\end{table*}

\subsection{Comparison Design and Scoring}
\label{app:gavel-scoring}
\rebuttal{The five sources are the clinician-authored timeline and the text-only and multimodal timelines from GLM-5.2 and DeepSeek V3.2. All 40 cases contribute to the primary comparison graph. Table~\ref{tab:gavel-schedule} gives the valid pairwise schedules for both judge families. They differ because not all pairwise outputs were valid under both judges; the analysis is not restricted to their intersection.}

\begin{table*}[t]
\centering
\caption{\rebuttal{Valid GAVEL comparisons by source pair and judge. Each row counts cases with a valid judgment for that pair. All 40 cases contribute to each overall graph. MM and TO denote multimodal and text-only reconstruction.}}
\label{tab:gavel-schedule}
\small
\setlength{\tabcolsep}{6pt}
\rebuttal{\begin{tabular}{lrr}
\toprule
Source pair & GLM-5.2 judge & MiMo judge\\
\midrule
DeepSeek MM vs. DeepSeek TO & 39 & 39\\
DeepSeek MM vs. GLM-5.2 MM & 36 & 36\\
DeepSeek MM vs. GLM-5.2 TO & 36 & 35\\
DeepSeek TO vs. GLM-5.2 MM & 37 & 37\\
DeepSeek TO vs. GLM-5.2 TO & 37 & 35\\
GLM-5.2 MM vs. GLM-5.2 TO & 37 & 35\\
Clinician vs. DeepSeek MM & 39 & 36\\
Clinician vs. DeepSeek TO & 40 & 38\\
Clinician vs. GLM-5.2 MM & 36 & 33\\
Clinician vs. GLM-5.2 TO & 37 & 36\\
\midrule
Total comparisons & 374 & 360\\
\bottomrule
\end{tabular}}
\end{table*}

\rebuttal{A decisive finding is charged to the source judged incorrect. Table~\ref{tab:gavel-weights} gives the study-defined weights. Verdicts \texttt{BOTH}, \texttt{NEITHER}, and \texttt{UNCLEAR} receive no points.}

\begin{table*}[t]
\centering
\caption{\rebuttal{Mapping from GAVEL findings to scored error categories.}}
\label{tab:gavel-weights}
\footnotesize
\setlength{\tabcolsep}{4pt}
\rebuttal{\begin{tabular}{@{}p{0.18\textwidth}p{0.69\textwidth}r@{}}
\toprule
Scored error & Mapping rule & Weight\\
\midrule
Over-annotation & Unsupported one-sided event; charged to the source containing it. & 3.0\\
False duplicate & Extra copy rejected by the judge. & 2.0\\
Wrong time & Decisive \texttt{TIMING} finding. & 2.0\\
Wrong value & Decisive \texttt{VALUE} finding. & 2.0\\
Missed positive & Supported one-sided positive event; charged to the omitting source. & 1.0\\
Missed negative & Supported one-sided negative finding; charged to the omitting source. & 0.5\\
Missed recurrence & Repeated occurrence supported by the record but omitted by the other source. & 0.5\\
\bottomrule
\end{tabular}}
\end{table*}

Let $E_{A,g}$ and $E_{B,g}$ be weighted errors charged to A and B in game $g$. A earns the cost charged to B, giving
\[
s_{A,g}=\begin{cases}
E_{B,g}/(E_{A,g}+E_{B,g}),&E_{A,g}+E_{B,g}>0,\\
1/2,&E_{A,g}+E_{B,g}=0.
\end{cases}
\]
\rebuttal{An all-nondecisive game remains a draw. Bradley--Terry strengths are fitted to game-level shares and displayed on the centered rating scale}
\[
R_m=1500+\frac{400}{\log 10}\left(\log\theta_m-\frac1M\sum_{j=1}^M\log\theta_j\right).
\]
\rebuttal{The pairwise model is $\Pr(m\succ n)=\theta_m/(\theta_m+\theta_n)$. Only rating differences within an analysis carry meaning. Marginal intervals summarize performance against the full graph and are not equivalence tests between two selected sources. Intervals use 2,000 case-level bootstrap resamples (seed 20260904), retaining all games associated with each selected case.}

\rebuttal{Direct multimodal-versus-text-only comparisons use only games in which those two variants met. We report the multimodal source's share of severity-weighted points, with 50\% denoting parity, and 2,000 case-bootstrap intervals. This point share is distinct from the proportion of cases won. Case wins, losses, nonempty ties, and empty games are reported separately in Appendix~\ref{apd:gavel_results}.}

\subsection{\rebuttal{Judge Robustness and Compute}}
\label{app:gavel-robustness-protocol}
\rebuttal{The primary judge is GLM-5.2. The MiMo-V2.6-Pro-RL analysis changes judge family while retaining the prompt, evidence inputs, weights, and scoring procedure. MiMo is not used for reconstruction or reference annotation. Each judge is analyzed on its own valid schedule (Table~\ref{tab:gavel-schedule}); this is not a matched-intersection estimate of judge-family effects.}

\rebuttal{Two further GLM-5.2 runs reverse A/B positions and add occurrence identifiers to the candidate rows, respectively. In the UID-bearing run, legacy annotation identifiers are produced deterministically from the case identifier, row index, event text, and timestamp using BLAKE2s, with salted rehashing to resolve collisions. These identifiers do not create a shared gold occurrence alignment across independently constructed timelines. The rerun tests input serialization, not an oracle UID-matching evaluator.}

\rebuttal{Both judges run locally on H200 GPUs without proprietary API calls. The reported full-cohort wall-clock times are approximately 12 hours for GLM-5.2 and 20 hours for MiMo. These are observed execution times, not standardized GPU-hour or monetary-cost comparisons. The available records do not specify a common GPU count for normalizing the two runs.}

\subsection{Clinical Validation Protocol}
\label{app:gavel-validation-protocol}
\rebuttal{Manual review uses the primary GLM-5.2 findings. Ten cases are sampled from the 36 with valid judgments for all three pairings among clinician-authored, GLM-5.2 multimodal, and DeepSeek V3.2 multimodal timelines. The other four cases remain in the overall evaluation; they are outside this common three-pairing frame. Case sets unable to supply five value findings per pairing are rejected and redrawn. This feasibility constraint means the final ten-case sample is conditioned on rare-finding availability.}

\rebuttal{Within the selected cases, 50 findings per pairing are sampled using the allocation in Table~\ref{tab:gavel-manual-sampling}. The 150 findings are split into three packets balanced across pairings and strata. The primary reviewer scores all packets, and the second reviewer independently scores one 50-finding packet. Reviewers see the verdict and record whether GAVEL handled the finding correctly, with free-text comments. This is a confirmation study, not verdict-blinded adjudication.}

\begin{table*}[t]
\centering
\caption{\rebuttal{Manual-review allocation. Rare types are oversampled; decisive and unclear timing findings form separate strata.}}
\label{tab:gavel-manual-sampling}
\small
\setlength{\tabcolsep}{8pt}
\rebuttal{\begin{tabular}{lrr}
\toprule
Stratum & Per pairing & Total\\
\midrule
One-sided & 18 & 54\\
\texttt{TIMING}, decisive & 15 & 45\\
\texttt{TIMING}, \texttt{UNCLEAR} & 5 & 15\\
\texttt{DUPLICATE} & 7 & 21\\
\texttt{VALUE} & 5 & 15\\
\midrule
All strata & 50 & 150\\
\bottomrule
\end{tabular}}
\end{table*}

\rebuttal{Raw confirmation uses the primary reviewer's 150 decisions, without double-counting the shared packet. The standardized rate weights each stratum's confirmation rate by its frequency in the eligible finding corpus. Its interval uses 20,000 stratified bootstrap resamples (seed 20260910). Inter-reviewer agreement uses the shared 50 findings. Standardization corrects the deliberate finding-type oversampling; it does not establish performance on unreviewed cases or recall of discrepancies never emitted by GAVEL.}

%%%%%%%%%%%%%%%%%%%%%%%%

% \FloatBarrier
\section{Evaluation of Textual Time Series}
\label{apd:tts_evaluation}
\rebuttal{We evaluate reconstructed textual time series through reference-event recovery, temporal ordering, and timestamp agreement.} \rebuttal{Temporal metrics use only matched events, so they must be interpreted alongside event match rate. Reconstruction UIDs are not shared by the clinician inventory or prior-method baselines and do not replace semantic matching in this evaluation.}

\subsection{Event Match Rate}
\label{apd:event-match-rate}
\rebuttal{Following \citet{wang2025large,noroozizadeh2026reconstructing}, we use recursive one-to-one best matching with PubMedBERT cosine distance. At each iteration, all remaining reference--prediction pairs are considered and the minimum-distance pair is selected. Pairs meeting the primary distance threshold of 0.1 are retained. Exact ties are broken by reference index and then prediction index. Algorithm~\ref{alg:recursive_match} makes these steps explicit.}

\[
\mathrm{MatchRate}=\frac{N_{\mathrm{matched}}}{N_{\mathrm{ref}}},
\]
where $N_{\mathrm{matched}}$ is the number of reference events with a matched prediction and $N_{\mathrm{ref}}$ is the total number of reference events.
\rebuttal{The common 40-case reference inventory contains 9,147 events. Pairing is based on event descriptions; timestamps are used to evaluate temporal agreement after pairing.}

\begin{algorithm2e*}[t]
\footnotesize
\caption{\rebuttal{Recursive best matching with explicit threshold filtering and index-based tie-breaking}}
\label{alg:recursive_match}
\SetAlgoLined
\SetKwInOut{Input}{Input}
\SetKwInOut{Output}{Output}
\SetKwProg{Fn}{Function}{:}{}
\ifshowrebuttalcolor\color{rebuttalblue}\fi
\Input{Reference list $R$, predicted list $P$, cosine-distance threshold $\tau=0.1$}
\Output{One-to-one semantic matches}
\Fn{\texttt{MatchEvents}($R,P,\tau$)}{
  \If{$R$ or $P$ is empty}{\Return{[\,]}\;}
  Compute $d_{ij}$ for every remaining reference--prediction pair\;
  Select $(i^*,j^*)$ minimizing $(d_{ij},i,j)$ in lexicographic order\;
  \If{$d_{i^*j^*}>\tau$}{\Return{[\,]}\;}
  $r^*\gets R[i^*]$; $p^*\gets P[j^*]$\;
  Remove $R[i^*]$ and $P[j^*]$ from their lists\;
  \Return{$[(r^*,p^*)]+\texttt{MatchEvents}(R,P,\tau)$}\;
}
\end{algorithm2e*}

\subsection{Temporal Concordance}
\label{apd:temporal-concordance}
\rebuttal{We measure temporal ordering agreement with the concordance index. Let $t_i^{\mathrm{ref}}$ and $t_i^{\mathrm{pred}}$ be the reference and predicted times for matched event $i$. The c-index is}
\[
\mathrm{c}=\frac{1}{N}\sum_{\substack{i<j\\t_i^{\mathrm{ref}}\ne t_j^{\mathrm{ref}}\\t_i^{\mathrm{pred}}\ne t_j^{\mathrm{pred}}}}
\mathds{1}\!\left\{(t_i^{\mathrm{ref}}-t_j^{\mathrm{ref}})(t_i^{\mathrm{pred}}-t_j^{\mathrm{pred}})>0\right\},
\]
where $N$ is the number of comparable pairs. Higher values indicate better preservation of the reference ordering.

\subsection{Time Discrepancy and AULTC}
\label{apd:aultc}
\rebuttal{Following \citet{noroozizadeh2026reconstructing}, the absolute time error for matched event $i$ is $\Delta t_i=|t_i^{\mathrm{pred}}-t_i^{\mathrm{ref}}|$. We transform it to $x_i=\log(1+\Delta t_i)$ and form the empirical cumulative distribution}
\[
F(x)=\frac{1}{k}\sum_{i=1}^{k}\mathds{1}\{x_i\le x\},
\]
where $k$ is the number of matched events pooled across the evaluated cases.

\rebuttal{AULTC summarizes the normalized area under this log-time discrepancy curve. The normalization limit is fixed at $S_{\max}=43{,}800$ hours (five 365-day years), giving a log-time limit of $\log(1+43{,}800)$. It is the same for every method, model, ablation, matching threshold, and bootstrap replicate; it is not the maximum observed error and is not estimated separately for each method. Larger AULTC values indicate closer timestamp agreement.}

\subsection{\rebuttal{Bootstrap Intervals and Paired Comparisons}}
\label{app:paired-bootstrap}
\rebuttal{The marginal intervals in Table~\ref{tab:main-results} and Table~\ref{tab:model-performance} use 200 case-level bootstrap resamples. Each resample includes all evaluated events from a selected case, retaining their dependence. Paired comparisons retain cases with valid outputs in both conditions, draw those cases with replacement 10,000 times, and use the same draw for both conditions. We recompute the metric difference within every replicate and report the 2.5th and 97.5th percentiles. All five GLM-5.2 paired-ablation comparisons use the same 40 cases (seed 1).}

\rebuttal{A paired interval excluding zero supports a difference for that metric and comparison. An interval containing zero does not establish equivalence. These analyses keep the semantic-matching threshold fixed; they do not force the variants to recover the same events. A higher temporal score may therefore accompany lower event recovery. Table~\ref{tab:best_ablation} reports all five GLM-5.2 paired comparisons, and Table~\ref{tab:paired-multimodal-summary} reports the cross-backbone intervals that exclude zero.}

\subsection{\rebuttal{Matching-Order and Exact-Duplicate Sensitivity}}
\label{app:matching-sensitivity}
\rebuttal{For GLM-5.2 and DeepSeek V3.2, we independently permuted reference and predicted event order 1,000 times per case and reran the same matcher. Event match rate was unchanged in every permutation. Across the tested models, the largest absolute model-level changes were 0.0051 in concordance and 0.0072 in AULTC (Table~\ref{tab:matching-permutation}). Exact-distance ties can change which repeated occurrence is paired and therefore change its associated time, even when the number of matches is unchanged.}

\begin{table}[t]
\centering
\caption{\rebuttal{Largest absolute model-level change across the 1,000 order permutations for the two tested backbones.}}
\label{tab:matching-permutation}
\small
\rebuttal{\begin{tabular}{lr}
\toprule
Metric & Maximum absolute change\\
\midrule
Event match rate & 0.0000\\
Concordance & 0.0051\\
AULTC & 0.0072\\
\bottomrule
\end{tabular}}
\end{table}

\rebuttal{A separate sensitivity analysis collapses exact duplicate mention--timestamp rows before matching while retaining repeated mentions at different times. Table~\ref{tab:duplicate-sensitivity} reports the resulting scores for the three tested multimodal backbones. GLM-5.2 has the highest point estimate on all three metrics in this analysis. This tests sensitivity to exact duplicate entries; it does not verify that every semantic match identifies the correct source occurrence.}

\begin{table*}[t]
\centering
\caption{\rebuttal{Exact-duplicate sensitivity. Identical mention--timestamp rows are collapsed before semantic matching; distinct-time recurrences are retained. Entries are estimates with 95\% intervals.}}
\label{tab:duplicate-sensitivity}
\small
\setlength{\tabcolsep}{7pt}
\rebuttal{\begin{tabular}{lccc}
\toprule
Multimodal backbone & Event match rate & Concordance & AULTC\\
\midrule
GLM-5.2 & 0.757 (0.731--0.792) & 0.851 (0.830--0.874) & 0.793 (0.761--0.826)\\
DeepSeek V3.2 & 0.624 (0.597--0.646) & 0.767 (0.739--0.788) & 0.759 (0.731--0.775)\\
Qwen3.5-397B & 0.625 (0.596--0.652) & 0.780 (0.743--0.821) & 0.765 (0.732--0.795)\\
\bottomrule
\end{tabular}}
\end{table*}

\subsection{\rebuttal{Semantic-Matching Threshold Sweeps}}
\label{app:threshold-sweeps}
\rebuttal{We vary the semantic distance threshold from 0.01 to 0.50 in steps of 0.01. Figure~\ref{fig:main_tts_sensitivity} plots concordance and AULTC against the resulting match rate for the main methods. Figure~\ref{fig:ablation_threshold_sweeps} gives the corresponding three-backbone ablation plots. These are evaluation-threshold sweeps, not sensitivity analyses of the retrieval-score threshold, \texttt{known} flag, or whether an event time changed.}

%%%%%%%%%%%%%%%%%%%%%%%%

% \FloatBarrier
\section{Direct Comparison with TKW2}
\label{apd:tkw2-comparison}
\begin{table*}[t]
\centering
\setlength{\tabcolsep}{5pt}
\renewcommand{\arraystretch}{1.15}
\caption{Methodological comparison between TKW2 and the present UID-preserving
framework. The current method extends the same text-primary multimodal
formulation but adds occurrence identity, explicit provenance, and
source-grounded adjudication.}
\label{tab:tkw2-design-comparison}
\footnotesize
\begin{tabular}
{@{}p{0.14\textwidth}p{0.39\textwidth}p{0.39\textwidth}@{}}
\toprule
\textbf{Dimension} & \textbf{TKW2} & \textbf{Present framework} \\
\midrule
Event unit
& Standalone event descriptions propagated as text strings; central and
non-central events are constructed in separate stages.
& Source occurrences tagged once, each with a persistent UID and character
span; repeated or similar mentions remain distinct. \\

Text-only \mbox{initialization}
& Central events and pairwise offsets form a temporal scaffold; non-central
events are subsequently attached to that scaffold.
& A complete UID-indexed event inventory is assigned point times, bounds,
textual-explicitness flags, and contextual UID links before structured evidence
is introduced. \\

Structured \mbox{retrieval}
& Direct event-to-row retrieval using MedTE embeddings and top-$k$ cosine
similarity; structured rows are supplied to two calibration stages.
& UID-conditioned LLM queries retrieve patient-specific event-series summaries;
Qwen3 embedding and reranking are followed by expansion to timestamped source
rows. \\

Timeline revision
& Structured evidence calibrates the central scaffold and later the assembled
full timeline.
& One joint pass revises the complete timeline while requiring every UID and
original mention to be preserved; up to three complete alternatives may be
returned. \\

Traceability
& No persistent identifier links a final event to one source occurrence or to
the row that changed its time.
& Explicit chain from source span to UID, initial estimate, query, structured
source row, and revised time. \\

Evaluation
& Event match rate, concordance, AULTC, scaffold ablations, and a structured-data
gap analysis.
& The same reference-based metrics, component ablations for identity and
evidence linkage, \rebuttal{and source-grounded adjudication of event-and-time timelines; the primary GAVEL analysis omits UIDs.} \\
\bottomrule
\end{tabular}
\vspace{-3mm}
\end{table*}

\citet{kumar2026text} introduced \emph{Text Knows What, Tables Know When}
(TKW2), the closest predecessor to the present framework. TKW2 reconstructs a
central-event scaffold from narrative text, calibrates that scaffold with
retrieved structured EHR rows, attaches non-central events, and performs a
second structured-data refinement over the assembled timeline. Its main result
was that structured evidence often improved temporal localization while having
little effect on event recovery.

The present framework retains the same text-primary view of reconstruction but
changes the unit carried through the pipeline from a mutable event string to a
source-grounded occurrence. Table~\ref{tab:tkw2-design-comparison} summarizes
the resulting methodological differences. \rebuttal{Persistent UIDs and character spans preserve the identity of each extracted source occurrence within reconstruction. Source-row links expose the retrieved timing evidence, and the separate GAVEL evaluator reports disagreements against the record.}

TKW2 established the value of structured rows as partial temporal evidence: its
multimodal variants generally left event match rate stable while improving one
or both temporal metrics. The present results remain model-dependent, so they
do not support a universal numerical improvement from multimodality. The main
advance over TKW2 is instead the ability to hold the target occurrence fixed,
trace each revision to source evidence, test those mechanisms through targeted
ablations, and adjudicate disagreements without treating one timeline as an
infallible reference.

\rebuttal{Table~\ref{tab:main-results} compares both pipelines using DeepSeek V3.2 on the same cohort and matching threshold. Multimodal event match rate is 0.482 for TKW2 and 0.617 for the present pipeline, an absolute difference of 0.135 (28.0\% relative). The corresponding concordance values are 0.774 and 0.762, and AULTC values are 0.771 and 0.754. Higher event recovery therefore accompanies lower temporal point estimates on the respective matched sets. This is a comparison of complete pipelines, not an isolated estimate of the effect of UIDs. The GLM-5.2 no-UID ablation in Table~\ref{tab:best_ablation} tests identifier removal within the present pipeline; its event-recovery interval includes zero.}

%%%%%%%%%%%%%%%%%%%%%%%%

% \FloatBarrier
\section{Extended GAVEL Results}
\label{apd:gavel_results}
% Included under:
% \section{Extended GAVEL Results}
% \label{apd:gavel_results}
%
% Requires: booktabs, makecell, multirow, adjustbox.

\begin{table*}[p]
\centering
\caption{\rebuttal{Manual confirmation and primary GLM-5.2 GAVEL results.} \textbf{(a)} Shared-set inter-reviewer agreement. \textbf{(b)} Primary-review confirmation by validation stratum. \textbf{(c)} Bradley--Terry ratings. \textbf{(d)} Controlled within-backbone comparisons. Intervals are 95\% confidence intervals; MM and UM denote multimodal and unimodal reconstruction.}
\label{tab:gavel-validation}
\label{tab:gavel-overview}
\label{tab:gavel-ratings}
\label{tab:gavel-controlled}
\fontsize{7.0}{7.7}\selectfont
\setlength{\tabcolsep}{2.6pt}
\renewcommand{\arraystretch}{0.96}

\begin{minipage}[t]{0.405\textwidth}
\vspace{0pt}
\centering
\textbf{(a) Shared-set inter-reviewer agreement ($n=50$)}\\[-1pt]
\begin{tabular}{lrrr}
\toprule
& \multicolumn{2}{c}{Reviewer 2} & \\
\cmidrule(lr){2-3}
Reviewer 1 & Uphold & Reject & Total \\
\midrule
Uphold & 42 & 1 & 43 \\
Reject & 4 & 3 & 7 \\
\midrule
Total & 46 & 4 & 50 \\
\bottomrule
\end{tabular}

\vspace{1.5pt}
\raggedright
Raw agreement: 45/50, 90.0\% [78.6, 95.7].
Cohen's $\kappa=0.494$; PABAK $=0.800$.
\end{minipage}
\hfill
\begin{minipage}[t]{0.565\textwidth}
\vspace{0pt}
\centering
\textbf{(b) Manual confirmation by validation stratum}\\[-1pt]
\begin{tabular}{lrrrr}
\toprule
Stratum & $n$ & Upheld & Rate (\%) & Corpus (\%) \\
\midrule
\texttt{TIMING}, decisive & 45 & 34 & 75.6 & 51.2 \\
\texttt{TIMING}, \texttt{UNCLEAR} & 15 & 9 & 60.0 & 14.8 \\
One-sided & 54 & 50 & 92.6 & 27.2 \\
\texttt{DUPLICATE} & 21 & 17 & 81.0 & 4.6 \\
\texttt{VALUE} & 15 & 12 & 80.0 & 2.2 \\
\midrule
Primary review & 150 & 122 & 81.3 & 100.0 \\
\bottomrule
\end{tabular}

\vspace{1.5pt}
\raggedright
Primary raw rate: 81.3\% [74.3, 86.8]; corpus-standardized rate:
78.2\% [70.3, 85.8] from 20,000 stratified bootstrap resamples.
Reviewer 2: 46/50, 92.0\% [81.2, 96.8].
\end{minipage}

% Explicit separation between panels (b) and (c).
\par\vspace{8pt}

\begin{minipage}[t]{0.985\textwidth}
\vspace{0pt}
\centering
\textbf{(c) Analysis scope and Bradley--Terry ratings}\\[-1pt]
5 sources; 40 cases; 374 games; 13,321 findings; 10,924 decisive (82.0\%);
2,397 non-scoring (18.0\%); 26 draws, including 19 games with no decisive finding.\\[1.5pt]
\begin{tabular}{lrrr}
\toprule
Source & Rating & 95\% interval & Mean share \\
\midrule
Clinician-authored & 1555 & 1505--1608 & 59.8\% \\
GLM-5.2 MM & 1541 & 1513--1570 & 57.2\% \\
GLM-5.2 UM & 1482 & 1449--1515 & 46.8\% \\
DeepSeek V3.2 MM & 1475 & 1441--1507 & 45.7\% \\
DeepSeek V3.2 UM & 1447 & 1418--1476 & 40.6\% \\
\bottomrule
\end{tabular}
\end{minipage}

% Explicit separation between panels (c) and (d).
\par\vspace{8pt}

\begin{minipage}[t]{0.985\textwidth}
\vspace{0pt}
\centering
\textbf{(d) Controlled multimodal--unimodal comparisons}\\[-1pt]
\begin{tabular}{lrrrrrrr}
\toprule
Backbone & Cases & Empty & Points & MM share & 95\% interval & W--L--T & Separated \\
\midrule
GLM-5.2 & 37 & 7 & 1,882 & 79.6\% & 66.5--90.0 & 21--8--1 & Yes \\
DeepSeek V3.2 & 39 & 12 & 828 & 56.5\% & 39.1--74.4 & 16--10--1 & No \\
\bottomrule
\end{tabular}

\vspace{1.5pt}
\raggedright
MM share is the percentage of weighted points earned by the multimodal source;
50\% denotes parity. Empty games contain no decisive finding and remain draws.
\end{minipage}

\vspace{7pt}

\caption{\rebuttal{Primary GLM-5.2 judge: error profiles, weight sensitivity, and annotation-protocol strata.} \textbf{(a)} Raw totals and errors charged per matchup. \textbf{(b)} Bradley--Terry ratings under alternative weights. \textbf{(c)} Clinician-annotation protocol strata. Percentages may not sum to 100 because of rounding.}
\label{tab:gavel-error-totals}
\label{tab:gavel-error-profile}
\label{tab:gavel-weight-sensitivity}
\label{tab:gavel-annotation-protocol}
\fontsize{6.65}{7.25}\selectfont
\setlength{\tabcolsep}{1.7pt}
\renewcommand{\arraystretch}{0.94}

\textbf{(a) Decisive error distribution and source-specific rates per matchup}\\[-1pt]
\begin{tabular}{lrrrrrrrrr}
\toprule
Source or summary & Over & False dup. & Wrong time & Wrong value & Missed pos. & Missed neg. & Missed recur. & Total & Games \\
\midrule
Raw $n$ & 130 & 559 & 6,572 & 271 & 2,955 & 405 & 32 & 10,924 & -- \\
Share of decisive (\%) & 1.2 & 5.1 & 60.2 & 2.5 & 27.1 & 3.7 & 0.3 & 100.0 & -- \\
\midrule
DeepSeek V3.2 MM & 0.15 & 0.62 & 8.51 & 0.13 & 3.02 & 0.39 & 0.05 & 12.86 & 150 \\
DeepSeek V3.2 UM & 0.08 & 0.92 & 10.21 & 0.20 & 4.12 & 0.84 & 0.01 & 16.38 & 153 \\
GLM-5.2 MM & 0.17 & 0.67 & 8.34 & 0.27 & 2.55 & 0.25 & 0.07 & 12.33 & 146 \\
GLM-5.2 UM & 0.22 & 0.64 & 11.90 & 0.23 & 2.97 & 0.29 & 0.07 & 16.31 & 147 \\
Clinician-authored & 0.25 & 0.88 & 5.05 & 0.97 & 6.99 & 0.92 & 0.02 & 15.07 & 152 \\
\midrule
Average source & 0.17 & 0.75 & 8.79 & 0.36 & 3.95 & 0.54 & 0.04 & 14.60 & 149.6 \\
\bottomrule
\end{tabular}

\vspace{5pt}

\begin{minipage}[t]{0.655\textwidth}
\vspace{0pt}
\centering
\textbf{(b) Sensitivity to error weights}\\[-1pt]
\setlength{\tabcolsep}{1.35pt}
\begin{tabular}{lrrrrrr}
\toprule
Scheme & Clin. & GLM MM & GLM UM & DS MM & DS UM & Draws \\
\midrule
Prespecified
& \makecell{\textbf{1555}\\{[1505,1608]}}
& \makecell{1541\\{[1513,1570]}}
& \makecell{1482\\{[1449,1515]}}
& \makecell{1475\\{[1441,1507]}}
& \makecell{1447\\{[1418,1476]}} & 26 \\
Equal
& \makecell{\textbf{1542}\\{[1490,1596]}}
& \makecell{1536\\{[1510,1562]}}
& \makecell{1484\\{[1452,1515]}}
& \makecell{1481\\{[1447,1514]}}
& \makecell{1457\\{[1428,1485]}} & 32 \\
Timing only
& \makecell{\textbf{1608}\\{[1559,1661]}}
& \makecell{1553\\{[1520,1590]}}
& \makecell{1480\\{[1441,1519]}}
& \makecell{1446\\{[1411,1479]}}
& \makecell{1413\\{[1377,1446]}} & 39 \\
Recall only
& \makecell{1465\\{[1405,1521]}}
& \makecell{1487\\{[1454,1522]}}
& \makecell{1519\\{[1485,1553]}}
& \makecell{\textbf{1520}\\{[1482,1558]}}
& \makecell{1509\\{[1464,1556]}} & 107 \\
\bottomrule
\end{tabular}
\end{minipage}
\hfill
\begin{minipage}[t]{0.325\textwidth}
\vspace{0pt}
\centering
\textbf{(c) Clinician-annotation protocol strata}\\
\setlength{\tabcolsep}{1.25pt}
\begin{tabular}{lrrrr}
\toprule
& \multicolumn{2}{c}{Unaided} & \multicolumn{2}{c}{LLM-assisted} \\
\cmidrule(lr){2-3}\cmidrule(lr){4-5}
Source & Pen. & Rating & Pen. & Rating \\
\midrule
Clinician & 34.21 & 1417 & 10.71 & 1719 \\
GLM MM & 20.36 & 1527 & 23.01 & 1556 \\
GLM UM & 20.09 & 1528 & 37.30 & 1434 \\
DS MM & 16.09 & 1538 & 28.14 & 1399 \\
DS UM & 18.78 & 1489 & 35.58 & 1392 \\
\bottomrule
\end{tabular}

\vspace{1.5pt}
\raggedright
Penalty is weighted cost per matchup; lower is better. Ratings are fitted and
centered independently within each subset and are not comparable across rating columns.
\end{minipage}

\end{table*}

\rebuttal{Tables~\ref{tab:gavel-validation} and \ref{tab:gavel-error-profile} collect the primary GLM-5.2 judge's manual-review, comparison, and error-profile results. Table~\ref{tab:gavel-new-sensitivity} adds the MiMo, position-swapped, and UID-bearing analyses, and Table~\ref{tab:gavel-mimo-profile} gives MiMo's error counts. The two judges differ in the number of valid comparisons and in how many findings they emit; their raw finding totals are not treated as directly comparable error rates.}

\subsection{Manual Validation}
\label{app:gavel-manual-validation}

\rebuttal{Reviewers saw GAVEL's verdict and recorded whether its handling of the finding was correct, including whether the finding should have been raised or left unresolved. The study estimates confirmation of reported findings; it does not independently search complete timelines for missed discrepancies.}
The primary reviewer scored a stratified sample of 150 findings, and a second
reviewer independently scored 50 of those findings. On the shared set, the
reviewers agreed on 45 of 50 decisions (90.0\%; 95\% CI, 78.6--95.7), with
Cohen's $\kappa=0.494$ and prevalence-adjusted bias-adjusted $\kappa=0.800$
(Table~\ref{tab:gavel-validation}\textbf{(a)}). Because both reviewers usually upheld the
judge, the labels were highly imbalanced; raw agreement is therefore the
primary inter-reviewer measure.

All five inter-reviewer disagreements concerned whether GAVEL should have
emitted or resolved a finding, rather than which candidate timeline was
better supported. They involved timing differences within the judge's
tolerance, an abstention for which the evidence was arguably sufficient,
and a typographical discrepancy in which one timeline reproduced the note
and the other corrected it. The primary reviewer generally enforced the
prompt's emission and abstention rules, whereas the second reviewer focused
on whether the stated verdict was substantively correct. Four of the five
disagreements arose from one case, accounting for four of its 11 shared
findings, compared with one disagreement among the remaining 39 findings.

The primary reviewer upheld 122 of 150 findings (81.3\%; 95\% CI,
74.3--86.8). The sample deliberately oversampled uncommon finding types, so the
raw rate does not estimate performance under the observed corpus distribution.
Standardizing the five validation strata to their corpus shares yielded a
confirmation rate of 78.2\% (95\% CI, 70.3--85.8; 20,000 stratified bootstrap
resamples). One-sided findings were upheld most often (92.6\%), whereas
\texttt{TIMING} findings assigned \texttt{UNCLEAR} were the most difficult
(60.0\%; Table~\ref{tab:gavel-validation}\textbf{(b)}). The second reviewer upheld 46 of
the 50 shared findings (92.0\%; 95\% CI, 81.2--96.8), but this overlapping
subset is not pooled with the primary review.

\subsection{Five-Source Comparison}
\label{app:gavel-five-source-results}

All 40 cases contributed to 374 available pairwise games, comprising 13,321
findings; 10,924 (82.0\%) were decisive. Under the prespecified severity
weights, the clinician-authored and GLM-5.2 multimodal timelines had the two
highest Bradley--Terry point estimates (Table~\ref{tab:gavel-ratings}\textbf{(c)}). Their
marginal intervals overlap and summarize performance against the complete
comparison graph rather than a direct test between the two sources.

The within-backbone comparisons in Table~\ref{tab:gavel-controlled}\textbf{(d)} more
directly isolate access to structured evidence. GLM-5.2 multimodal earned
79.6\% of weighted points against GLM-5.2 unimodal (95\% CI, 66.5--90.0),
separating from parity. DeepSeek V3.2 multimodal earned 56.5\% against its
unimodal counterpart (39.1--74.4), which did not separate from 50\%.

\subsection{Error Profiles and Robustness}
\label{app:gavel-error-results}

Wrong timing accounted for 6,572 decisive findings (60.2\%), and missed
positive events accounted for 2,955 (27.1\%); together they comprised 87.2\%
of decisive findings (Table~\ref{tab:gavel-error-profile}\textbf{(a)}). Both multimodal model
sources had fewer errors per matchup than their unimodal counterparts. The
clinician-authored source had the lowest wrong-time rate but the highest
missed-positive and wrong-value rates.

The Bradley--Terry order was unchanged under equal and timing-only weights
(Table~\ref{tab:gavel-weight-sensitivity}\textbf{(b)}). Recall-only scoring changed the
order because it removed the dominant timing category; this is a diagnostic
extreme rather than an alternative primary specification.

Table~\ref{tab:gavel-annotation-protocol}\textbf{(c)} stratifies the comparison by the
protocol used to create the clinician-authored timeline. Its weighted penalty
fell from 34.21 per matchup in the unaided subset to 10.71 in the LLM-assisted
subset, \rebuttal{These are different subsets, not a before--after change in the same cases.} The subsets contain different cases and are
confounded with data source, and their ratings were fitted and centered
separately; this contrast is descriptive rather than causal.

% Both composite tables are kept in one float so they remain on the same page.

\subsection{\rebuttal{Independent Judge, Candidate Position, and UID Inputs}}
\label{app:gavel-added-robustness}
\rebuttal{MiMo-V2.6-Pro-RL produced 33,890 findings from 360 valid comparisons over all 40 cases; 28,067 were decisive. The primary GLM-5.2 run produced 13,321 findings from 374 comparisons. Table~\ref{tab:gavel-new-sensitivity}\textbf{(a)} summarizes their analysis scope. The valid schedules differ (Table~\ref{tab:gavel-schedule}); no common-schedule restriction was applied.}

\rebuttal{The two judges exchange the top two overall point estimates: clinician-authored ranks first under GLM-5.2 and second under MiMo, while GLM-5.2 multimodal moves from second to first. The remaining point-estimate order is unchanged (Table~\ref{tab:gavel-new-sensitivity}\textbf{(b)}). These separately centered marginal ratings do not establish clinician--model equivalence or a direct test of one source against another.}

\rebuttal{The direct within-backbone conclusions agree. Under MiMo, GLM-5.2 multimodal earns 75.2\% of weighted points against text-only (95\% CI, 62.2--86.3), while DeepSeek multimodal earns 57.7\% (39.4--76.6). The corresponding primary-judge shares are 79.6\% and 56.5\%. Both GLM-5.2 intervals exclude parity, and both DeepSeek intervals include it (Table~\ref{tab:gavel-new-sensitivity}\textbf{(c)}). The figures describe weighted-point shares, not case-win rates.}

\rebuttal{Under the GLM-5.2 judge, reversing candidate positions preserves the five-source rating order, as does adding UID-bearing input rows. The maximum absolute shift in the UID-bearing ratings is nine points. These analyses support stability of the aggregate ordering under the tested changes; they neither eliminate all judge bias nor establish finding-level agreement. The primary manual-confirmation results apply to the original GLM-5.2 findings and were not repeated for MiMo.}

\begin{table*}[p]
\centering
\caption{\rebuttal{Additional GAVEL analyses. \textbf{(a)} Scope under each judge family. \textbf{(b)} Ratings with 95\% intervals under the primary, position-swapped, UID-bearing, and MiMo analyses. \textbf{(c)} Direct within-backbone results. MM and TO denote multimodal and text-only. All intervals use case-level resampling; ratings are centered separately.}}
\label{tab:gavel-new-sensitivity}
\footnotesize
\setlength{\tabcolsep}{4pt}
\renewcommand{\arraystretch}{1.06}
\rebuttal{\textbf{(a) Judge-analysis scope}}\par\vspace{3pt}
\rebuttal{\begin{tabular}{lrrrrrr}
\toprule
Judge & Cases & Comparisons & Findings & Decisive & Draws & Empty\\
\midrule
GLM-5.2 & 40 & 374 & 13,321 & 10,924 & 26 & 19\\
MiMo-V2.6-Pro-RL & 40 & 360 & 33,890 & 28,067 & 26 & 21\\
\bottomrule
\end{tabular}}
\par\vspace{11pt}
\rebuttal{\textbf{(b) Five-source ratings across robustness conditions}}\par\vspace{3pt}
\rebuttal{\begin{tabular}{lcccc}
\toprule
Source & GLM primary & A/B reversed & UID-bearing & MiMo judge\\
\midrule
Clinician-authored & \makecell{1555\\(1505--1608)} & \makecell{1570\\(1517--1623)} & \makecell{1547\\(1495--1599)} & \makecell{1523\\(1477--1570)}\\[3pt]
GLM-5.2 MM & \makecell{1541\\(1513--1570)} & \makecell{1549\\(1523--1577)} & \makecell{1536\\(1510--1562)} & \makecell{1546\\(1520--1575)}\\[3pt]
GLM-5.2 TO & \makecell{1482\\(1449--1515)} & \makecell{1482\\(1452--1511)} & \makecell{1491\\(1462--1518)} & \makecell{1494\\(1461--1526)}\\[3pt]
DeepSeek V3.2 MM & \makecell{1475\\(1441--1507)} & \makecell{1473\\(1436--1508)} & \makecell{1472\\(1437--1505)} & \makecell{1483\\(1452--1513)}\\[3pt]
DeepSeek V3.2 TO & \makecell{1447\\(1418--1476)} & \makecell{1427\\(1395--1455)} & \makecell{1454\\(1422--1481)} & \makecell{1454\\(1425--1483)}\\
\bottomrule
\end{tabular}}
\par\vspace{11pt}
\rebuttal{\textbf{(c) Direct multimodal-versus-text-only comparisons}}\par\vspace{3pt}
\rebuttal{\begin{tabular}{llrrrrcl}
\toprule
Judge & Backbone & Cases & Empty & Points & MM share & 95\% interval & W--L--T\\
\midrule
GLM-5.2 & GLM-5.2 & 37 & 7 & 1,882 & 79.6\% & 66.5--90.0 & 21--8--1\\
GLM-5.2 & DeepSeek V3.2 & 39 & 12 & 828 & 56.5\% & 39.1--74.4 & 16--10--1\\
MiMo & GLM-5.2 & 35 & 7 & 2,014 & 75.2\% & 62.2--86.3 & 21--6--1\\
MiMo & DeepSeek V3.2 & 39 & 12 & 1,496 & 57.7\% & 39.4--76.6 & 17--10--0\\
\bottomrule
\end{tabular}}
\par\vspace{4pt}
\begin{minipage}{0.96\textwidth}\footnotesize
\rebuttal{Empty games have no decisive findings and remain draws. W--L--T counts wins, losses, and ties among nonempty games; those three counts plus the empty count equal the number of cases. ``Points'' gives the combined weighted points in the direct comparison. Fifty percent is parity for the MM share.}
\end{minipage}
\end{table*}

\rebuttal{Table~\ref{tab:gavel-mimo-profile} records the MiMo finding counts and source-specific rates. It is a descriptive account of the second judge's outputs, not an independent clinical error annotation. Clinical confirmation has only been measured for the primary judge. Annotation-protocol reconstruction results are reported separately in Table~\ref{tab:annotation-protocol-reconstruction}.}

\begin{table*}[t]
\centering
\caption{\rebuttal{MiMo decisive findings: total counts and findings charged per source matchup. A matchup is one source against one opponent on one case. Rates divide by each source's valid matchup count. These outputs have not undergone the primary judge's manual-confirmation study.}}
\label{tab:gavel-mimo-profile}
\fontsize{7.2}{8.1}\selectfont
\setlength{\tabcolsep}{2.0pt}
\rebuttal{\begin{tabular}{lrrrrrrrrr}
\toprule
Source / total & Over & False dup. & Wrong time & Wrong value & Missed pos. & Missed neg. & Missed recur. & Total & Matchups\\
\midrule
Raw count & 217 & 3,700 & 13,125 & 790 & 8,698 & 1,279 & 258 & 28,067 & --\\
\midrule
DeepSeek V3.2 MM & 0.10 & 6.52 & 19.68 & 0.66 & 10.30 & 1.70 & 0.16 & 39.12 & 146\\
DeepSeek V3.2 TO & 0.13 & 5.46 & 18.28 & 0.87 & 11.55 & 2.26 & 0.89 & 39.44 & 149\\
GLM-5.2 MM & 0.26 & 1.79 & 16.71 & 0.34 & 7.50 & 0.72 & 0.16 & 27.48 & 141\\
GLM-5.2 TO & 0.37 & 8.35 & 20.40 & 0.33 & 11.77 & 1.72 & 0.19 & 43.13 & 141\\
Clinician-authored & 0.66 & 3.54 & 16.05 & 3.29 & 19.27 & 2.45 & 0.36 & 45.63 & 143\\
\bottomrule
\end{tabular}}
\end{table*}

 %%%%%%%%%%%%%%%%%%%%%%%%
 
% \FloatBarrier
\section{Extended Ablation Analyses}
\label{apd:baselines_ablations}
\paragraph{Ablation variants.}
\rebuttal{We evaluate six configurations for GLM-5.2, DeepSeek V3.2, and Qwen3.5-397B. \emph{Full multimodal} uses the complete pipeline. \emph{Text-only} is the initial timeline before structured evidence is introduced. \emph{No UID} removes identifiers supplied to joint revision while retaining the input event inventory, estimates, evidence, and backbone. \emph{No query generation} retrieves with the source mention. \emph{No source-row linkage} provides retrieved summary text without its mapped timestamped rows. \emph{No reranker} omits the reranking stage. Table~\ref{tab:model-performance} reports the absolute scores and marginal intervals.}

\begin{table*}[t]
\centering
\caption{Ablation performance across three reconstruction backbones at event-matching threshold $0.1$. Values are point estimates with 95\% confidence intervals from 200 case-level bootstrap resamples. Concordance and AULTC are computed on the events matched under each variant and should be interpreted together with event match rate.}
\label{tab:model-performance}
\small
\setlength{\tabcolsep}{5pt}
\renewcommand{\arraystretch}{1.08}
\begin{tabular}{llccc}
\toprule
\textbf{Model} & \textbf{Version} & \textbf{Event match rate} & \textbf{Concordance} & \textbf{AULTC} \\
\midrule
\multirow{6}{*}{GLM-5.2}
 & Full multimodal & 0.790 (0.750--0.826) & 0.802 (0.768--0.840) & 0.773 (0.736--0.812) \\
 & No reranker & 0.790 (0.750--0.825) & 0.802 (0.771--0.833) & 0.784 (0.745--0.821) \\
 & Unimodal & 0.790 (0.750--0.825) & 0.781 (0.760--0.817) & 0.758 (0.721--0.796) \\
 & No UID in timeline revision & 0.693 (0.576--0.786) & 0.794 (0.777--0.852) & 0.792 (0.745--0.832) \\
 & No query generation & 0.790 (0.750--0.825) & 0.793 (0.773--0.840) & 0.776 (0.736--0.812) \\
 & No source-row linkage & 0.790 (0.750--0.825) & 0.781 (0.757--0.828) & 0.755 (0.719--0.791) \\
\midrule
\multirow{6}{*}{DeepSeek V3.2}
 & Full multimodal & 0.617 (0.552--0.669) & 0.762 (0.746--0.815) & 0.754 (0.715--0.796) \\
 & No reranker & 0.465 (0.330--0.594) & 0.772 (0.738--0.814) & 0.790 (0.752--0.816) \\
 & Unimodal & 0.643 (0.589--0.699) & 0.751 (0.723--0.769) & 0.759 (0.721--0.803) \\
 & No UID in timeline revision & 0.504 (0.406--0.605) & 0.772 (0.754--0.830) & 0.760 (0.719--0.800) \\
 & No query generation & 0.393 (0.277--0.535) & 0.778 (0.741--0.825) & 0.783 (0.746--0.816) \\
 & No source-row linkage & 0.387 (0.295--0.522) & 0.752 (0.722--0.814) & 0.758 (0.713--0.798) \\
\midrule
\multirow{6}{*}{Qwen3.5-397B}
 & Full multimodal & 0.639 (0.531--0.725) & 0.750 (0.723--0.812) & 0.758 (0.736--0.783) \\
 & No reranker & 0.637 (0.529--0.724) & 0.752 (0.722--0.812) & 0.757 (0.734--0.782) \\
 & Unimodal & 0.750 (0.675--0.826) & 0.764 (0.732--0.814) & 0.739 (0.708--0.775) \\
 & No UID in timeline revision & 0.746 (0.667--0.807) & 0.777 (0.736--0.812) & 0.747 (0.717--0.781) \\
 & No query generation & 0.642 (0.532--0.731) & 0.752 (0.724--0.812) & 0.756 (0.733--0.783) \\
 & No source-row linkage & 0.688 (0.617--0.755) & 0.752 (0.716--0.803) & 0.747 (0.724--0.774) \\
\bottomrule
\end{tabular}
\end{table*}

\paragraph{\rebuttal{Paired GLM-5.2 comparisons.}}
\rebuttal{Table~\ref{tab:best_ablation} reports all five full-minus-comparator differences using 10,000 paired case-bootstrap resamples. AULTC is higher for the full pipeline than for text-only reconstruction by 0.0147 [0.0067, 0.0218], and higher than for no source-row linkage by 0.0181 [0.0112, 0.0254], with essentially unchanged event recovery. The concordance intervals include zero. Query-generation and reranking removals also have temporal intervals spanning zero.}

\rebuttal{Removing UIDs lowers GLM-5.2 event match rate from 0.790 to 0.693, but the paired difference is uncertain: 0.0969 [-0.0006, 0.2491]. We interpret this as a descriptive signal, not a resolved statistical effect. Its higher no-UID AULTC point estimate is computed on a smaller and different matched-event set, excludes unmatched occurrences, and has a paired interval spanning zero. It does not establish a better complete reconstruction.}

\paragraph{Other backbones.}
\rebuttal{DeepSeek V3.2 shows a recovery--timing trade-off. Full multimodal has the highest event match rate among its multimodal variants (0.617), while no reranking and no query generation have higher temporal point estimates but lower match rates (0.465 and 0.393). Its text-only match rate is 0.643. For Qwen3.5-397B, multimodal AULTC is higher than text-only AULTC (0.758 versus 0.739), but event recovery and concordance are lower. Removing UIDs raises its event-recovery and concordance point estimates while lowering AULTC. No component removal is uniformly harmful across backbones.}

\paragraph{\rebuttal{Paired multimodal effects across backbones.}}
\rebuttal{Table~\ref{tab:paired-multimodal-summary} lists the multimodal-minus-text-only intervals that exclude zero in the six-backbone analysis. All concordance intervals and the unlisted match-rate and AULTC intervals include zero. These comparisons hold the matching threshold fixed at 0.1; gains in timing must therefore be read together with changes in event recovery. In particular, Qwen3.5's AULTC increase accompanies a lower match rate.}

\begin{table*}[t]
\centering
\caption{\rebuttal{Paired multimodal-minus-text-only differences with 95\% intervals excluding zero, from 10,000 case-level resamples. This is the subset of intervals excluding zero, not a complete table of all tested effects. All concordance intervals and unlisted match-rate/AULTC intervals include zero.}}
\label{tab:paired-multimodal-summary}
\small
\setlength{\tabcolsep}{8pt}
\rebuttal{\begin{tabular}{llrc}
\toprule
Metric & Backbone & Difference & 95\% interval\\
\midrule
Event match rate & DeepSeek V3.2 & $-0.030$ & [$-0.066$, $-0.002$]\\
Event match rate & Qwen3.5-397B & $-0.110$ & [$-0.225$, $-0.017$]\\
Event match rate & MiniMax M2.7 & $-0.018$ & [$-0.050$, $-0.001$]\\
AULTC & GLM-5.2 & $+0.015$ & [$+0.007$, $+0.022$]\\
AULTC & Qwen3.5-397B & $+0.019$ & [$+0.001$, $+0.042$]\\
\bottomrule
\end{tabular}}
\end{table*}

\begin{figure*}[p]
\centering
\setlength{\tabcolsep}{0pt}
\renewcommand{\arraystretch}{1.0}
\begin{tabular}{@{}>{\centering\arraybackslash}m{0.10\textwidth}@{\hspace{0.015\textwidth}}m{0.435\textwidth}@{\hspace{0.015\textwidth}}m{0.435\textwidth}@{}}
 & \centering\textbf{Concordance} & \centering\arraybackslash\textbf{AULTC}\\[4pt]
\small\textbf{GLM-5.2} & \includegraphics[width=\linewidth]{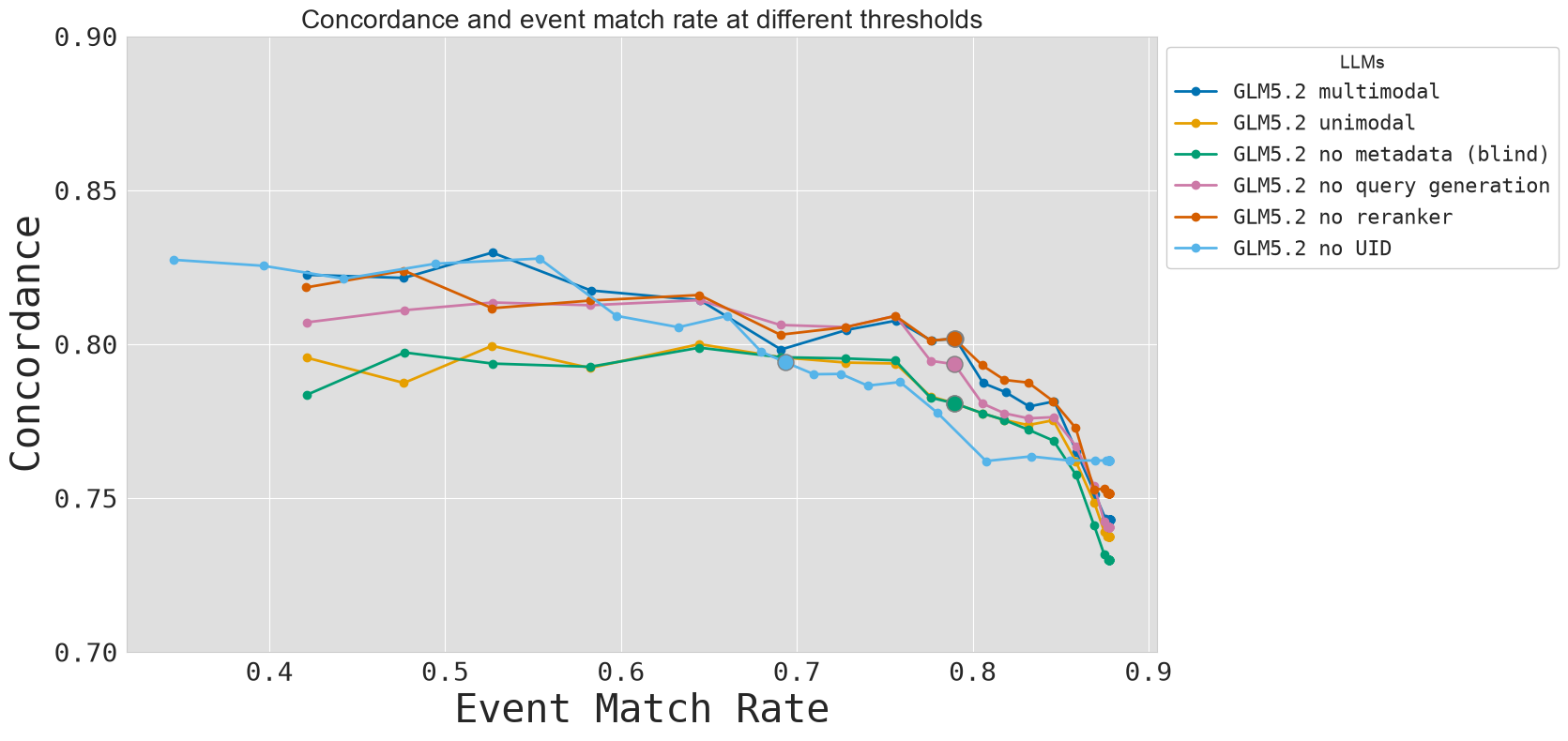} & \includegraphics[width=\linewidth]{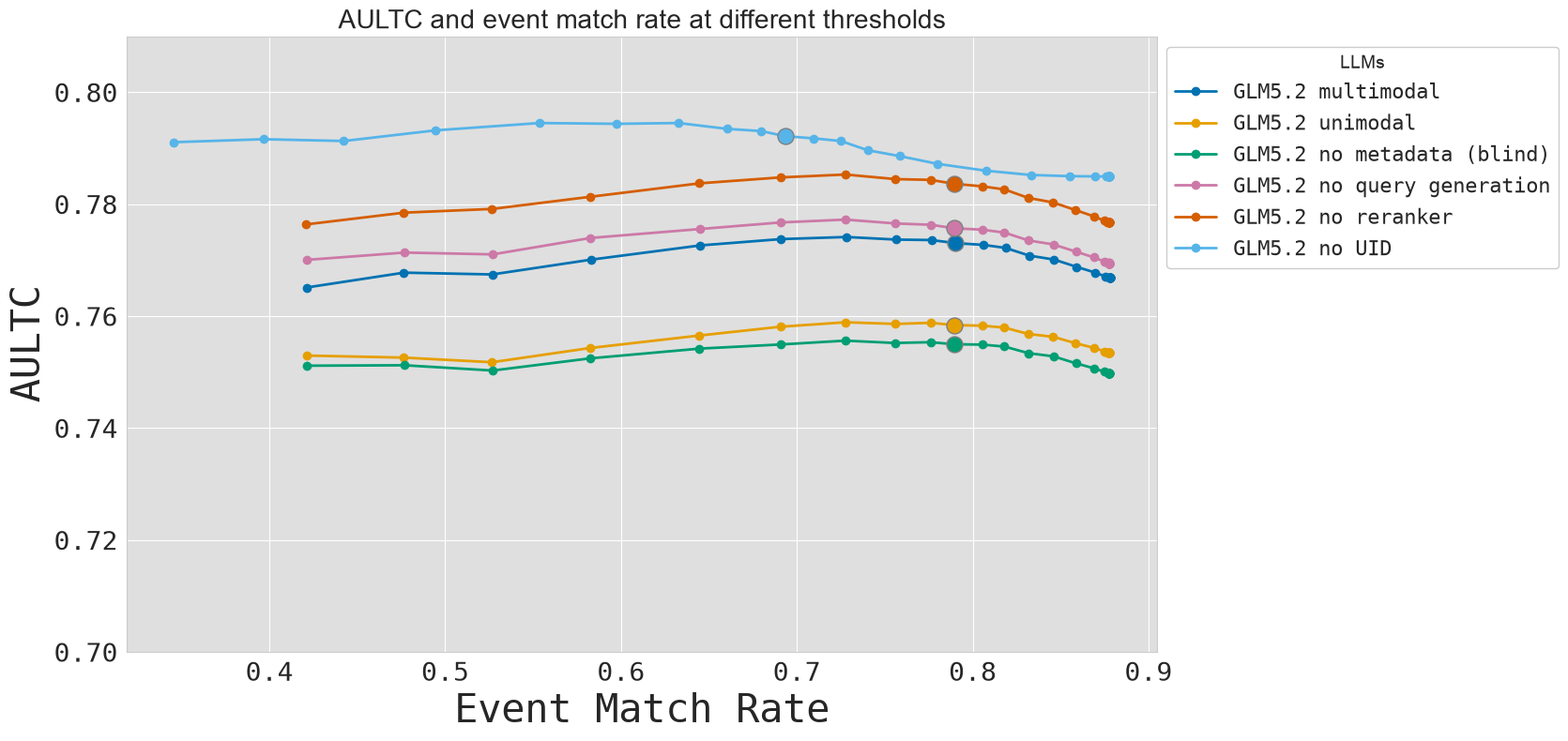}\\[6pt]
\small\textbf{DeepSeek V3.2} & \includegraphics[width=\linewidth]{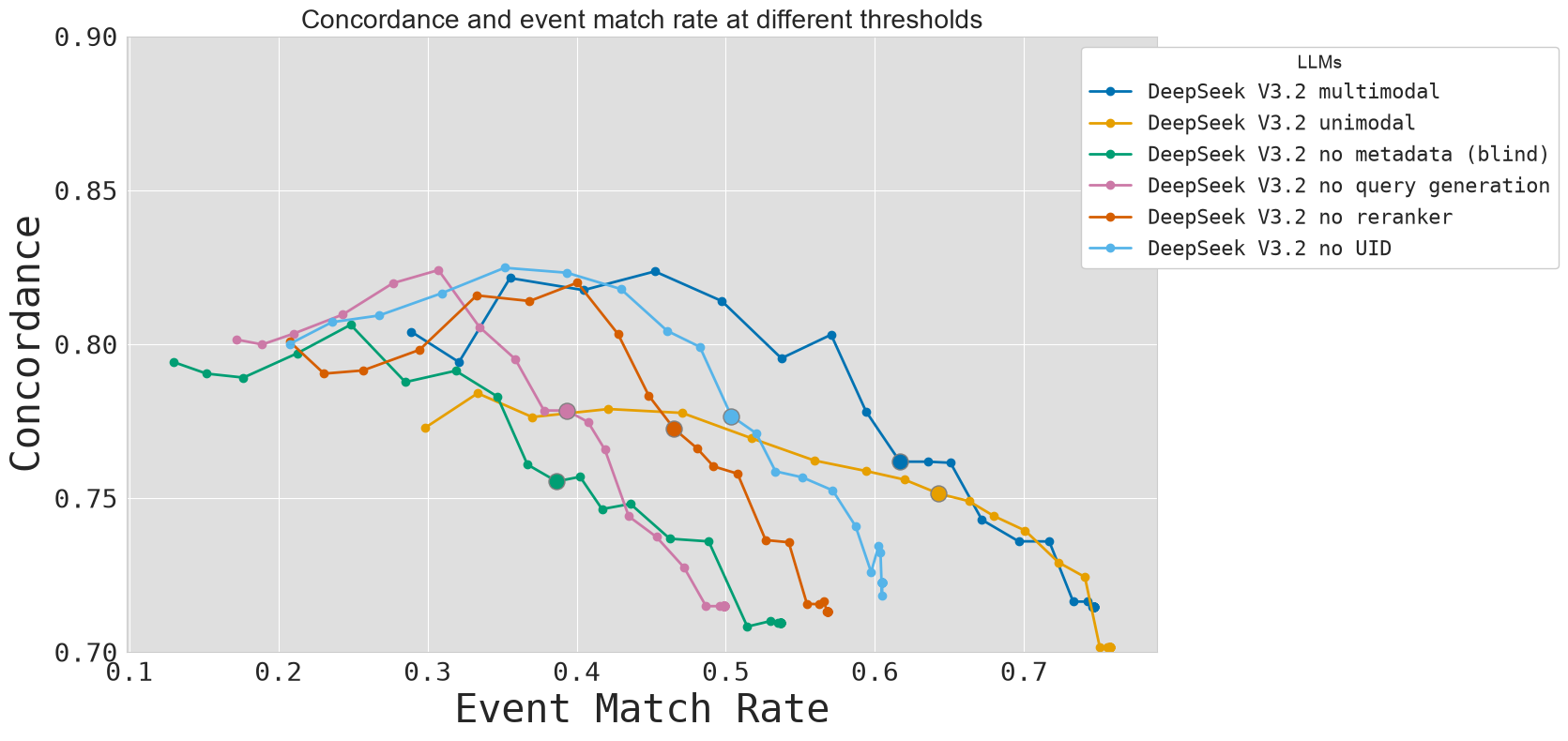} & \includegraphics[width=\linewidth]{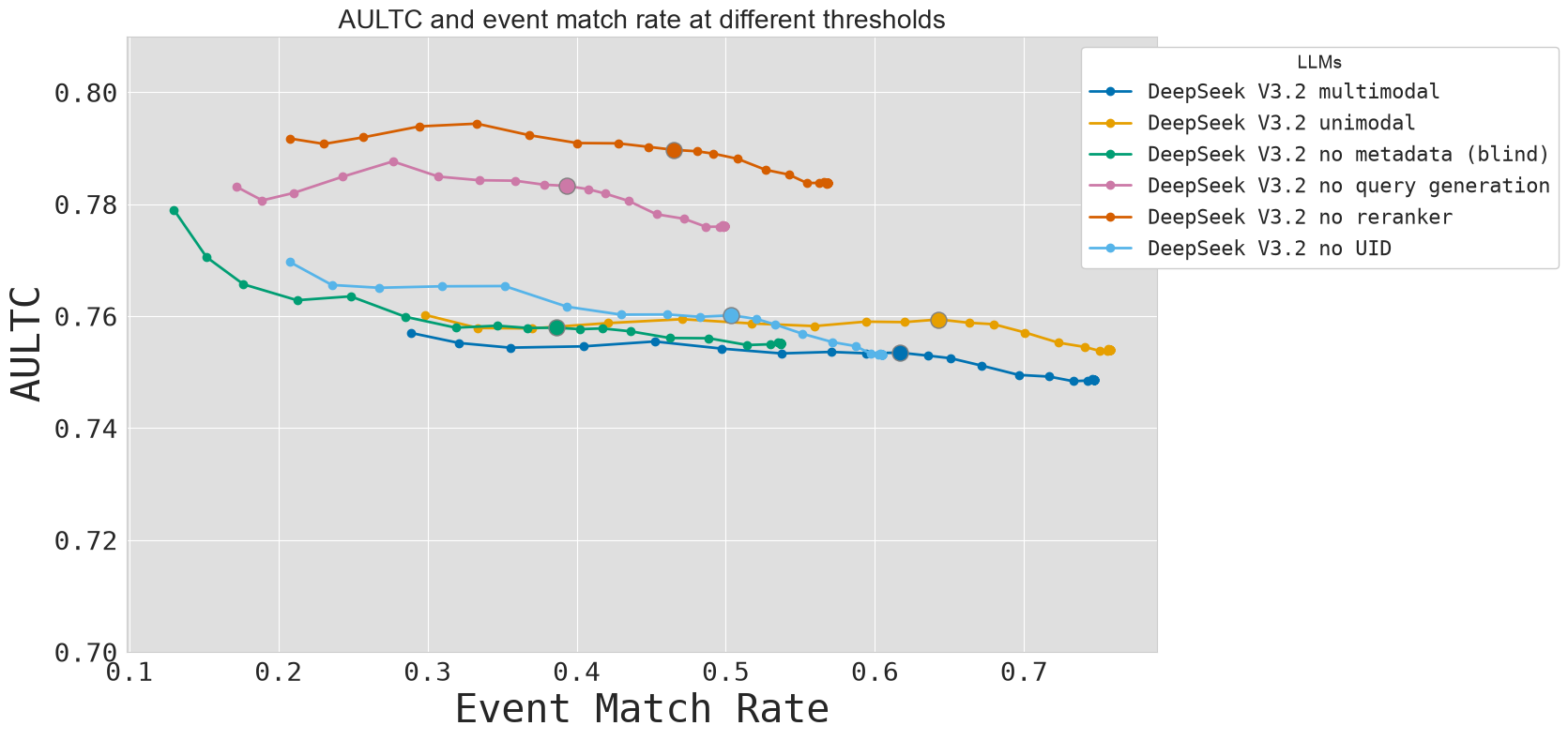}\\[6pt]
\small\textbf{Qwen3.5-397B} & \includegraphics[width=\linewidth]{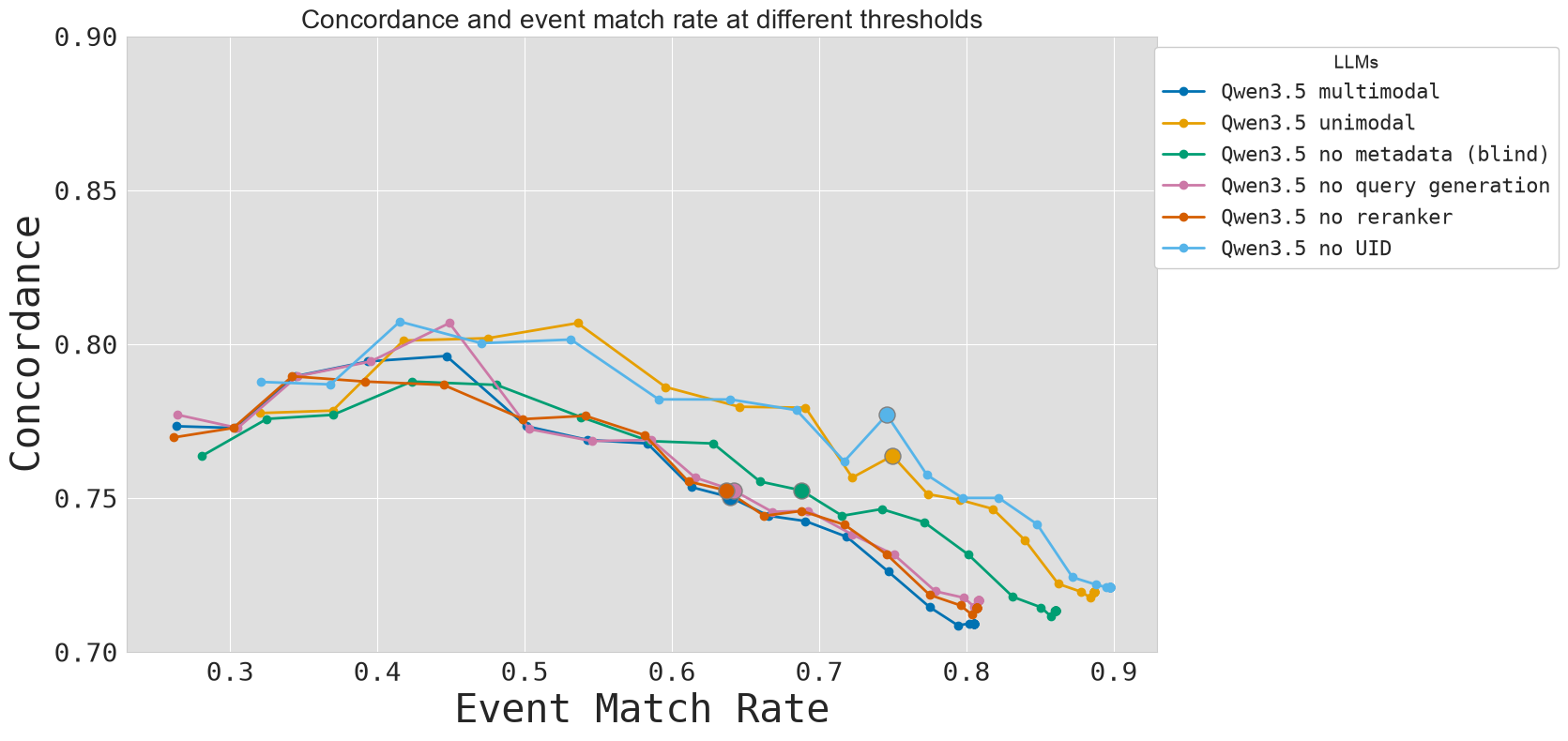} & \includegraphics[width=\linewidth]{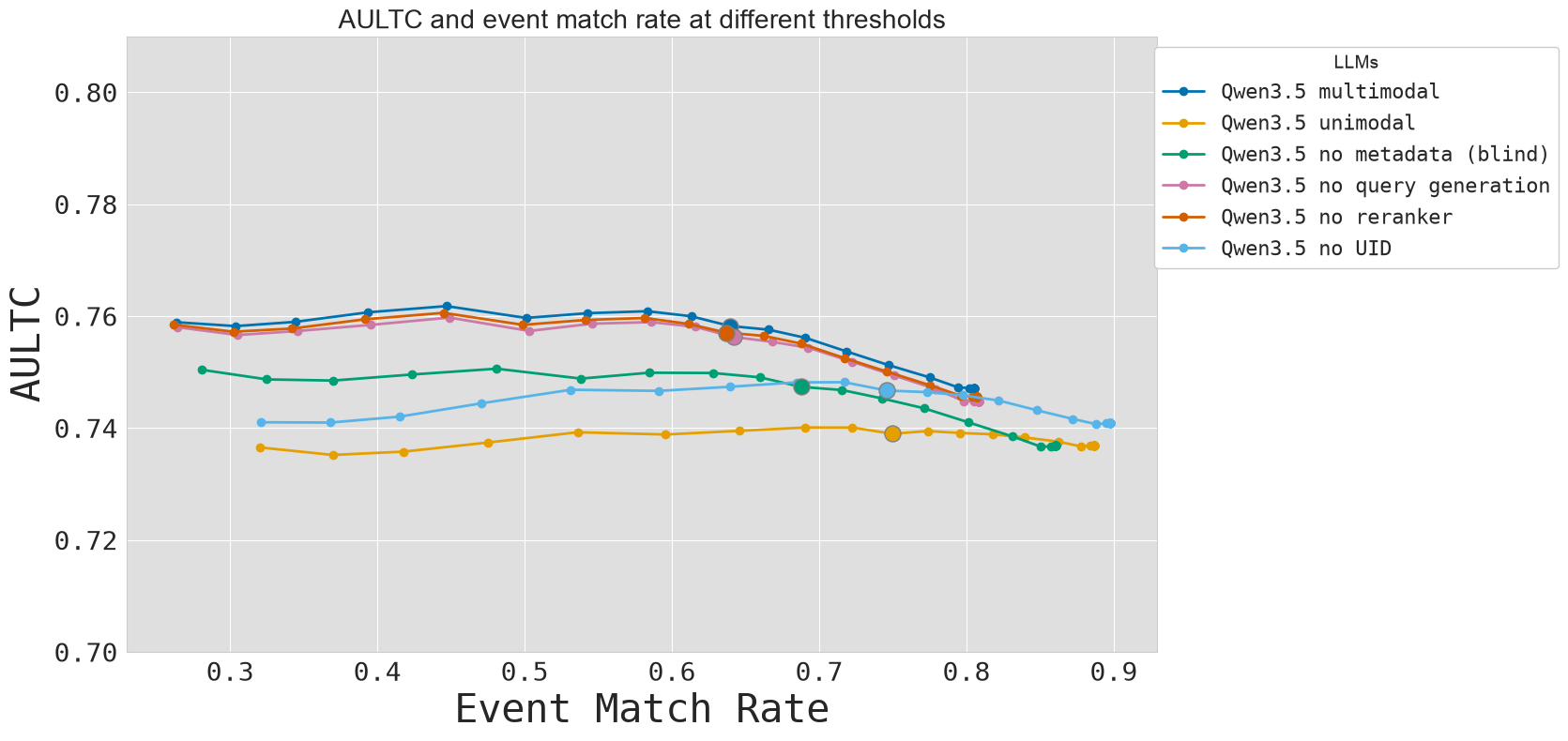}\\
\end{tabular}
\caption{\rebuttal{Threshold sweeps for the three reconstruction backbones used in the extended ablation analysis. Rows identify backbones; columns show concordance and AULTC. The semantic-matching threshold varies from 0.01 to 0.50 in steps of 0.01, and the horizontal axis gives the resulting event match rate. Enlarged markers denote the primary threshold of 0.1. ``No metadata (blind)'' in the original legends means no source-row linkage.}}
\label{fig:ablation_threshold_sweeps}
\end{figure*}

\paragraph{\rebuttal{Matching-threshold trends.}}
\rebuttal{Figure~\ref{fig:ablation_threshold_sweeps} varies the semantic-matching threshold from 0.01 to 0.50. At comparable match-rate ranges, the GLM-5.2 full AULTC curve generally exceeds the text-only and no-source-row curves; DeepSeek's higher-AULTC ablations cover fewer reference events. Concordance curves cross more frequently. These descriptive sweeps do not replace paired uncertainty at the prespecified threshold. In the original plot legends, ``no metadata (blind)'' denotes no source-row linkage and ``no UID'' denotes identifier removal during revision.}

%%%%%%%%%%%%%%%%%%%%%%%%
% \FloatBarrier
\section{Practical Use and Clinical Impact}
\label{apd:clinical-impact}
\paragraph{Time-sensitive cohorts and treatment windows.}
Retrospective cohort membership and treatment-window eligibility can change
when a symptom, diagnostic test, or intervention is moved across a temporal
boundary. The proposed framework preserves the narrative details needed to
define the clinical phenotype while using clinically matched EHR rows to refine
selected event times. In the synthetic case in
\figureref{fig:overview}, the reversal-agent occurrence retains the same UID
while a medication-administration record shifts its estimate from \(+1.5\) to
\(+6.5\) hours. This illustrates how structured evidence can alter temporal
placement without changing which occurrence is being timed.

\paragraph{\rebuttal{Occurrence time, information availability, and trajectory analysis.}}
\rebuttal{The reconstructed time estimates when a clinical event occurred. A prospective model can use that information only after it becomes available, which may be later because of result release, charting, or retrospective documentation. Leakage auditing therefore requires a separate availability timestamp and cannot rely on reconstructed occurrence time alone. Source provenance can help identify the relevant record and timestamp type. Retrospective temporal phenotyping and trajectory analyses can use the occurrence estimates while retaining narrative symptoms, functional status, pertinent negatives, and clinical interpretation.}

\paragraph{Clinician review and adjudication.}
The occurrence-level audit trail links each final timeline row to its narrative
span, generated retrieval queries, and timestamped structured evidence. A
reviewer can therefore inspect why a time changed rather than accepting a
revised timestamp without its provenance. \rebuttal{GAVEL separately compares event-and-time candidates, classifies their disagreements, and provides the narrative and structured evidence cited for each verdict. Its primary input omits reconstruction UIDs. Manual confirmation supports using these findings to guide review; missed-discrepancy recall and downstream clinical benefit remain unmeasured.}

The framework is intended to assist research data curation and
clinician-reviewed analysis. It has not been evaluated as an autonomous
decision-support system, and downstream studies are needed to determine whether
the reconstructed timelines improve prediction, trial screening, causal
analyses, or clinical decisions.

\end{document}